\pdfoutput=1
\documentclass[11pt]{article}
\usepackage[numbers]{natbib}
\usepackage{emnlp2021}
\usepackage{amsmath} 
\usepackage{times}
\usepackage{latexsym}
\usepackage{adjustbox}
\usepackage[T1]{fontenc}
\usepackage[utf8]{inputenc}
\usepackage{microtype}

\usepackage{caption}
\usepackage{subcaption}
\usepackage{hyperref}
\usepackage{makecell}
\usepackage{tabularx}
\usepackage{multirow} 
\usepackage{booktabs} 
\usepackage{comment}

\title{A Comprehensive Survey of Masked Faces: Recognition, Detection, and Unmasking}

\author{Mohamed Mahmoud$^{a,b}$, Mahmoud SalahEldin Kasem$^{a,b}$, Hyun-Soo Kang$^{a}$\\
  $^a$Chungbuk National University, $^b$Assiut University \\ }

\begin{document}
\maketitle
\begin{abstract}
Masked face recognition (MFR) has emerged as a critical domain in biometric identification, especially by the global COVID-19 pandemic, which introduced widespread face masks. This survey paper presents a comprehensive analysis of the challenges and advancements in recognising and detecting individuals with masked faces, which has seen innovative shifts due to the necessity of adapting to new societal norms. Advanced through deep learning techniques, MFR, along with Face Mask Recognition (FMR) and Face Unmasking (FU), represent significant areas of focus. These methods address unique challenges posed by obscured facial features, from fully to partially covered faces. Our comprehensive review delves into the various deep learning-based methodologies developed for MFR, FMR, and FU, highlighting their distinctive challenges and the solutions proposed to overcome them. Additionally, we explore benchmark datasets and evaluation metrics specifically tailored for assessing performance in MFR research. The survey also discusses the substantial obstacles still facing researchers in this field and proposes future directions for the ongoing development of more robust and effective masked face recognition systems. This paper serves as an invaluable resource for researchers and practitioners, offering insights into the evolving landscape of face recognition technologies in the face of global health crises and beyond.
\end{abstract}

\section{Introduction}

In recent years, integrating facial recognition systems across diverse sectors, such as security, healthcare, and human-computer interaction, has revolutionized identity verification and access control. Nevertheless, the widespread adoption of face masks in response to the global COVID-19 pandemic has introduced unprecedented challenges to the remarkable performance of conventional facial recognition technologies. The masking of facial features has spurred research initiatives in masked face recognition, prompting the application of innovative deep-learning techniques to address this novel challenge. This sets the stage for exploring advanced strategies in face recognition, particularly under challenging conditions involving small or partially obscured faces~\cite{zhang2020refineface}.

\begin{figure*}[htbp] 
  \centering
  \includegraphics[width=\linewidth]{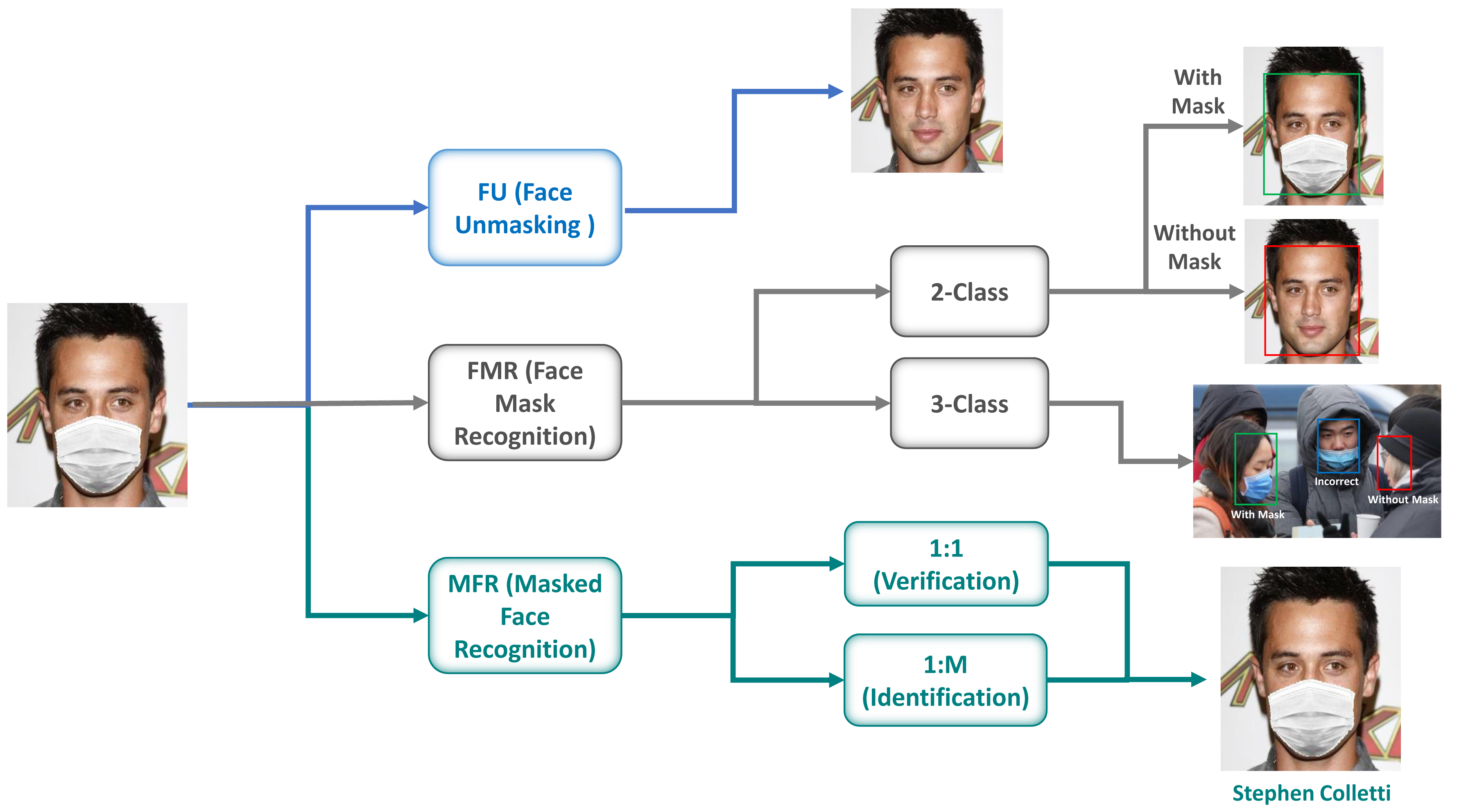}
  \caption{Illustration showcasing the tasks of Masked Face Recognition (MFR), Face Mask Recognition (FMR), and Face Unmasking (FU) with varied outputs for the same input.}
  \label{fig:1}
\end{figure*}

MFR poses a significant challenge in identifying and verifying individuals who wear face masks. This task is complex due to the partial occlusion and variations in appearance caused by facial coverings. Masks obscure critical facial features like the nose, mouth, and chin, and their diverse types, sizes, and colors add to the complexity. It's essential to distinguish between MFR and Face Mask Recognition (FMR). While FMR focuses on detecting mask presence, MFR aims to identify and verify individuals wearing masks. Additionally, Face Unmasking (FU) endeavors to remove facial coverings and restore a clear facial representation. Figure \ref{fig:1} visually summarizes our survey's essence, illustrating three distinct tasks—MFR, FMR, and FU—with different outputs for the same input, highlighting the task-specific outcomes.

Deep learning has emerged as a promising avenue for addressing the challenges of MFR. Algorithms can be trained to discern facial features, even when partially obscured by masks. Proposed MFR methodologies based on deep learning encompass holistic approaches, Mask Exclude-based approaches, and Mask Removal-based approaches. Holistic approaches employ deep learning models to discern features of entire faces, leveraging attention modules. Mask Exclude-based approaches train models to recognise features of the unmasked facial half, such as the eyes and head. Approaches based on Mask Removal leverage Generative Adversarial Networks (GANs) to create lifelike facial images from their masked counterparts, as demonstrated in methods \cite{mahmoud2023ganmasker, din2020novel}, facilitating subsequent recognition. Renowned for its exceptional proficiency beyond facial recognition tasks, deep learning has demonstrated remarkable success in many fields such as Optical Character Recognition (OCR) \cite{toiganbayeva2022kohtd, nurseitov2021handwritten, nurseitov2021classification, abdallah2020attention, kasem2023advancements}, network security \cite{mahmoud2022ae, zhu2023lkd, park2024conv3d, hnamte2023novel}, table detection \cite{kasem2022deep, abdallah2022tncr}, question answering \cite{abdallah2020automated,abdallah2024transformers,abdallah2023exploring,abdallah2023amurd, abdallah2023generator, abdallah2024arabicaqa}, and a diverse array of software applications \cite{hamada2021neural, kasem2024customer}. The adaptability of deep learning underscores its crucial role as a transformative technology in the realm of MFR. Although deep learning-based MFR methods have demonstrated state-of-the-art performance on various public benchmark datasets, numerous challenges persist that require resolution before widespread deployment in real-world applications becomes feasible.

\subsection{Challenges in MFR}
The advent of face masks amid the COVID-19 pandemic has posed formidable challenges to facial recognition systems, leading to a significant decline in their performance. The concealment of crucial facial features, including the nose, mouth, and chin, has triggered a cascade of obstacles, and addressing these challenges becomes imperative for advancing MFR technology.

\begin{itemize}
\item Scarcity of Datasets:
The scarcity of datasets tailored for masked face recognition constitutes a pivotal challenge. Training any deep learning model requires a robust dataset, yet the shortage of publicly available datasets featuring masked faces complicates the development of effective MFR methods. Researchers tackling this challenge often resort to creating synthetic datasets by introducing masks to existing public face datasets like CASIA-WebFace \cite{yi2014learning}, CelebA \cite{liu2015faceattributes}, and LFW \cite{huang2008labeled}. To simulate masked-unmasked pairs, popular methods involve using deep learning-based tools such as MaskTheFace \cite{anwar2020masked} or leveraging Generative Adversarial Networks like CycleGAN \cite{zhu2017unpaired}. Manual editing using image software, exemplified by the approach in \cite{din2020novel}, further supplements dataset generation efforts.

\item Dataset Bias:
In addition to the scarcity of publicly masked datasets, a prominent challenge lies in the bias inherent in existing benchmark datasets for MFR. Many widely used datasets exhibit a notable skew towards specific demographics, primarily favoring male and Caucasian or Asian individuals. This bias introduces a risk of developing MFR systems that may demonstrate reduced accuracy when applied to individuals from other demographic groups. To mitigate dataset bias in MFR, efforts should be directed towards creating more inclusive and representative benchmark datasets. This involves intentionally diversifying dataset populations to encompass a broader spectrum of demographics, including gender, ethnicity, and age. 

\item Occlusion Complexity:
The complexity introduced by facial occlusion, particularly the masking of the mouth, poses a significant hurdle to existing face recognition methods. The diverse sizes, colors, and types of masks exacerbate the challenge, impacting the training of models for various masked face tasks, including recognition, detection, and unmasking. Strategies to address this complexity vary by task. Recognition methods may employ attention models \cite{hu2018squeeze, woo2018cbam} that focus on the upper half of the face or exclusively train on this region. Another approach involves using face mask removal methods as a pre-step before recognition. In unmasking tasks, researchers may introduce a pre-stage to detect the mask area, as demonstrated by generating a binary mask map in the first stage in \cite{mahmoud2023ganmasker}. Training datasets are further diversified by incorporating various mask types, colors, and sizes to enhance model robustness. These nuanced approaches aim to unravel the intricacies posed by occlusions, ensuring the adaptability of masked face recognition methodologies.

\item Real-Time Performance:
Integrating masked face recognition into real-world scenarios poses intricate challenges, given the variability in lighting conditions, diverse camera angles, and environmental factors. Maintaining consistent performance amid these dynamic variables is a significant hurdle. Practical applicability across diverse settings necessitates real-time capabilities for MFR systems. However, the computational demands of deep learning-based MFR methods present a challenge, particularly when striving for real-time functionality on resource-constrained mobile devices. Addressing these real-time performance challenges involves a strategic optimization approach. Efforts focus on enhancing the efficiency of deep learning models without compromising accuracy.

\end{itemize}

\subsection{Applications of MFR}
MFR exhibits significant potential in numerous sectors, providing a secure and efficient means of identity verification in scenarios where individuals are wearing face masks. This potential translates into innovative solutions addressing contemporary challenges. This subsection explores the diverse applications where MFR can be leveraged, showcasing its adaptability and relevance.

\begin{itemize}
\item Security and Access Control:
Strengthening security measures to achieve precise identification, especially in scenarios involving individuals wearing masks. Seamlessly integrating with access control systems to guarantee secure entry across public and private spaces, including restricted areas like airports, government buildings, and data centers. Additionally, Implementing facial recognition-based door locks for both residential and office settings, enhancing home and workplace security. Enabling employee authentication protocols for secure entry into workplaces.

\item Public Safety:
MFR plays a crucial role in safeguarding public safety in crowded spaces. Integrated seamlessly with surveillance systems, MFR empowers law enforcement with enhanced monitoring and rapid response capabilities. This technology aids in identifying suspects and missing persons involved in criminal investigations, proactively detects suspicious activity in public areas, swiftly pinpoints individuals involved in disturbances, and strengthens security measures at events and gatherings. MFR's potential to enhance public safety and create a secure environment is undeniable.

\item Healthcare:
Ensuring secure access to medical facilities and patient records, along with verifying the identity of both patients and healthcare workers. Implementing contactless patient tracking to elevate healthcare services while simultaneously fortifying security and privacy within healthcare settings.

\item Retail and Customer Service:
Delivering tailored and efficient customer service by recognising individuals, even when their faces are partially obscured. Additionally, optimizing payment processes to elevate the overall shopping experience.

\item Human-Computer Interaction:
Facilitating secure and personalized interactions with user-authenticated devices while also improving the user experience across a spectrum of applications, including smartphones, computers, and smart home devices.

\item Workplace and Attendance Tracking:
Facilitating contactless attendance tracking for employees in workplace settings, thereby reinforcing security measures to grant access exclusively to authorized individuals in designated areas.

\item Education Institutions:
Overseeing and securing entry points in educational institutions to safeguard the well-being of students and staff. Streamlining attendance tracking in classrooms and campus facilities for enhanced efficiency.

\end{itemize}

By exploring these applications and more, it becomes evident that MFR has the potential to revolutionize diverse sectors, providing solutions that cater to the evolving needs of modern society.

Contributions of this survey encompass:
\begin{enumerate}
    \item An in-depth exploration of MFR, FMR, and FU within the framework of deep learning methodologies, highlighting the challenges inherent in identifying individuals with partially obscured facial features.
    \item A comprehensive exploration of evaluation metrics, benchmarking methods, and diverse applications of masked face recognition across security, healthcare, and human-computer interaction domains.
    \item A detailed analysis of critical datasets and preprocessing methodologies essential for training robust masked face recognition models.
    \item Tracing the evolutionary trajectory of face recognition within the deep learning paradigm, providing insights into the development of techniques tailored for identifying and verifying individuals under various degrees of facial occlusion.
\end{enumerate}

This investigation into masked face recognition, Face Mask Recognition, and Face Unmasking, grounded in the advancements of deep learning, aspires to furnish a foundational understanding for researchers. Serving as a roadmap, it delineates the current state-of-the-art methodologies and charts prospective avenues for continued research and development in this pivotal area.


\begin{figure*}[htbp] 
  \centering
  \includegraphics[width=\linewidth]{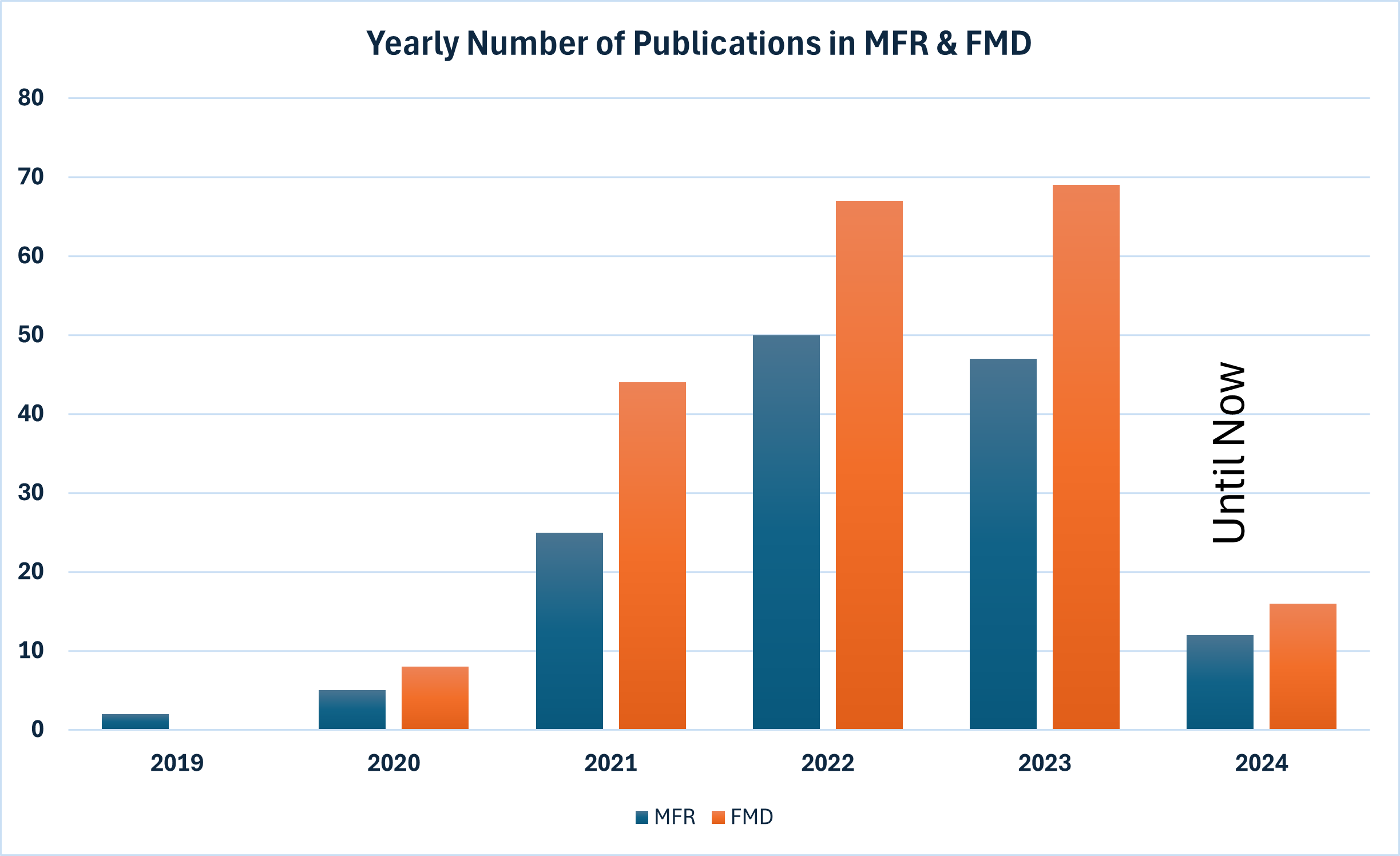}
  \caption{illustrates the evolving landscape of MFR and FMD studies from 2019 to 2024. The data was sourced from Scopus using keywords 'Masked face recognition' for MFR and 'Face mask detection', 'Face masks', and 'Mask detection' for FMD.}
  \label{fig:analysis}
\end{figure*}

\section{Related surveys}
In this section, we delve into previous surveys conducted in the fields of MFR and FMR, which serve as essential repositories of recent research and future directions. Despite the relatively short period following the onset of the COVID-19 pandemic, a considerable body of work and surveys has emerged. Figure \ref{fig:analysis} presents an analysis of related studies per year for both MFR and FMR from 2019 to 2024, with data sourced from Scopus. For MFR, the search was conducted using keywords such as "Masked face recognition" and "Masked Faces," while for FMR, keywords such as "Face mask detection," "Face masks," and "Mask detection" were utilized. These surveys offer invaluable insights into the evolution of methodologies, challenges encountered, and advancements made in tackling the intricacies associated with face masks. The objective of this subsection is to provide a succinct overview of select surveys in this domain, thereby situating the current study within the broader context of existing literature.

Face recognition under occlusion predates the COVID-19 era, indicating that Masked Face Recognition is not a novel field but rather a specialized and intricate subset of occluded face recognition. The complexity of masks, ranging from size and color to shape, adds layers of intricacy to MFR. Several surveys have explored partial face recognition, such as the work by Lahasan Badr et al \cite{lahasan2019survey}, which delves into strategies addressing three core challenges in face recognition systems: facial occlusion, single sample per subject (SSPS), and nuances in facial expressions. While offering insights into recent strategies to overcome these hurdles, this survey lacks recent updates and focus on Deep Learning methods, limiting its applicability to MFR challenges. Zhang Zhifeng et al \cite{zhang2020survey} address real-world complexities like facial expression variability and lighting inconsistencies alongside occlusion challenges. Despite being conducted during the COVID-19 pandemic, this survey overlooks masked face recognition and lacks comprehensive coverage of existing methods and empirical results. Similarly, Zeng Dan et al \cite{zeng2021survey} tackle the persistent challenge of identifying faces obscured by various occlusions, including medical masks. While categorizing modern and conventional techniques for recognising faces under occlusion, this survey lacks empirical results and comparative analyses of existing approaches addressing occlusion challenges.

Conversely, recent surveys have undertaken a comprehensive examination of MFR and FMR, with a focus on addressing the challenges encountered by face recognition and detection systems following the COVID-19 pandemic. Notably, Alzu’bi Ahmad et al. \cite{alzu2021masked} conducted an exhaustive survey on masked face recognition research, which has experienced significant growth and innovation in recent years. The study systematically explores a wide range of methodologies, techniques, and advancements in MFR, with a specific emphasis on deep learning approaches. Through a meticulous analysis of recent works, the survey aims to provide valuable insights into the progression of MFR systems. Furthermore, it discusses common benchmark datasets and evaluation metrics in MFR research, offering a robust framework for evaluating different approaches and highlighting challenges and promising research directions in the field.
Moreover, Wang Bingshu et al. \cite{wang2021survey} address the pressing need for AI techniques to detect masked faces amidst the COVID-19 pandemic. Their comprehensive analysis includes an examination of existing datasets and categorization of detection methods into conventional and neural network-based approaches. By summarizing recent benchmarking results and outlining future research directions, the survey aims to advance understanding and development in masked facial detection. Similarly, Nowrin Afsana et al. \cite{nowrin2021comprehensive} address the critical requirement for facemask detection algorithms in light of the global impact of the COVID-19 pandemic. Their study evaluates the performance of various object detection algorithms, particularly deep learning models, to provide insights into the effectiveness of facemask detection systems. Through a comprehensive analysis of datasets and performance comparisons among algorithms, the survey sheds light on current challenges and future research directions in this domain.


\section{Masked Face Datasets}
In the realm of MFR and FMR, the presence and quality of datasets play a pivotal role in shaping robust and accurate models. Datasets serve as the bedrock for unraveling the complexities associated with identifying individuals wearing face masks, making substantial strides in the progression of MFR methodologies. Furthermore, they establish the foundational framework for tasks such as face mask detection by creating paired masked-unmasked face datasets—integral components for training algorithms in face mask removal. This section extensively explores widely adopted standard benchmark datasets across various masked face tasks, encompassing MFR, FMR, and FU. To ensure a comprehensive overview, the section is bifurcated into two sub-sections based on the type of mask, distinguishing between real and synthetic masks. While real-world datasets offer heightened realism, they may be noisy and lack control. Conversely, while cleaner, synthetic datasets may not entirely capture real-world scenarios' intricacies.

\subsection{Real Mask Datasets}
\begin{figure*}[htbp]
    \begin{subfigure}[b]{0.3\textwidth}
         \includegraphics[width=\linewidth,height=.30\textheight]{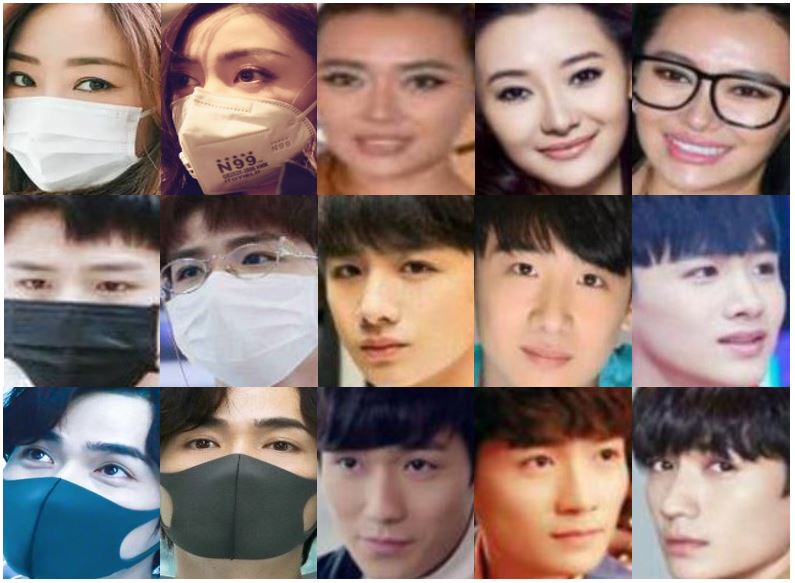}
         \subcaption{RMFD}\label{fig:a}
    \end{subfigure}
    \begin{subfigure}[b]{0.3\textwidth}
        \includegraphics[width=\linewidth,height=.30\textheight]{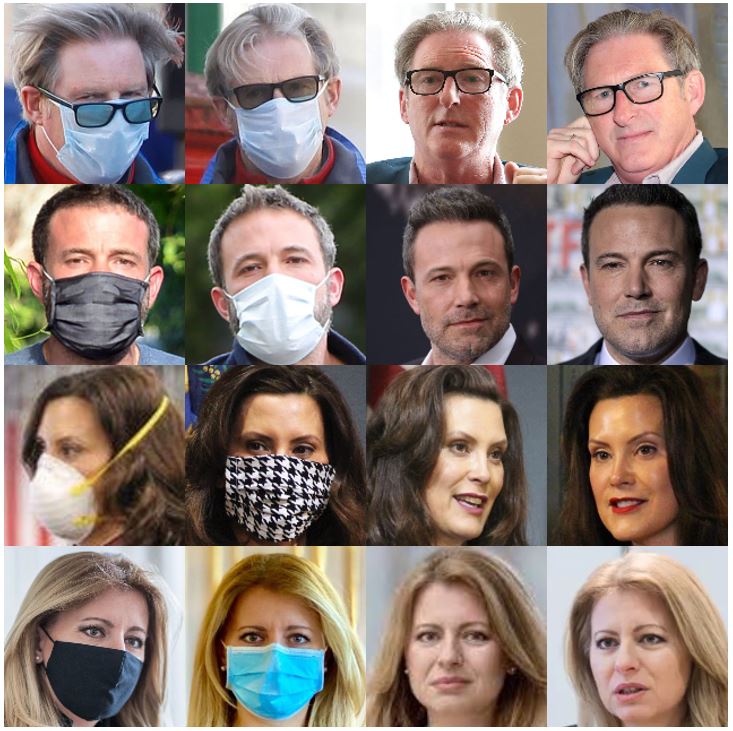}
        \subcaption{MFR2}\label{fig:b}
    \end{subfigure}
    \begin{subfigure}[b]{0.39\textwidth}
        \includegraphics[width=\linewidth,height=.30\textheight]{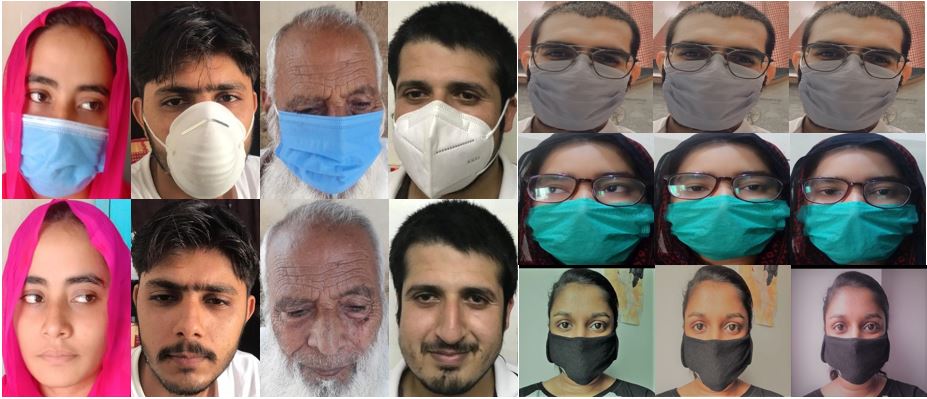}
        \subcaption{MDMFR}\label{fig:c}
    \end{subfigure}
    \caption{Samples of masked \& unmasked faces from the Real-Mask Masked Face Datasets used in Masked Face Recognition.}
    \label{fig:rmfd}
\end{figure*}

\begin{figure}[h] 
  \centering
  \includegraphics[width=0.5\textwidth]{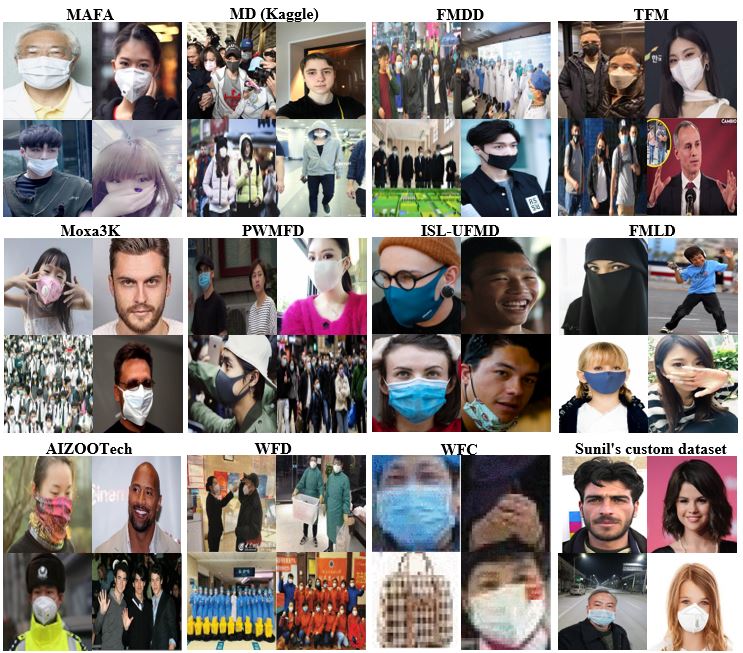}
  \caption{Samples from real masked face datasets used in Face Mask Recognition.}
  \label{fig:rmds}
\end{figure}

In the domain of MFR and FMR, datasets that incorporate genuine face masks provide crucial insight into the complexities posed by real-world scenarios. These datasets meticulously capture the intricacies of diverse face masks worn by individuals across various settings, ranging from public spaces and workplaces to social gatherings. The authenticity embedded in these masks significantly enhances the realism of the training process, allowing models to effectively adapt to the challenges presented by authentic face coverings. Notably, addressing a challenge highlighted in the discussion of limitations—the scarcity of real masked face datasets, especially those utilized in face mask removal tasks where pairing masked and unmasked faces is essential—this subsection delves into prominent benchmark datasets featuring real face masks. Through an exploration of these datasets, we illuminate their characteristics, applications, and significance in propelling advancements within the field of MFR methodologies. Table \ref{tab:real} offers an overview of the primary benchmark datasets utilized in both MFR and FMR, elucidating their essential attributes and utility across various applications. Furthermore, Figure \ref{fig:a} showcases sample images from the RMFD dataset, recognised as the most extensive dataset utilized for MFR. For comparison, Figure \ref{fig:b} displays samples from another significant dataset in the MFR domain, namely MFR2 \cite{anwar2020masked}. Meanwhile, Figure \ref{fig:rmds} illustrates sample images from datasets employed in FMR, offering visual insights into the various real mask datasets discussed.

\begin{table}[htbp]
\centering
\begin{adjustbox}{width={0.5\textwidth}}
  \begin{tabular}{c|c|c|c|c}
    \hline
    \textbf{Dataset} & \textbf{Size} & \textbf{Identities} & \textbf{Access} & \textbf{Year}\\
    \hline
    RMFRD \cite{Wang2020MaskedFaceRecognition} & \makecell{ 5000\\90000} & 525 & Public & 2020\\
    MFR2 \cite{anwar2020masked} & 269 & 53 & Public & 2020\\
    MASR-REC \cite{geng2020masked} & 11,615 & 1,004 & Private & 2020\\
    FMDD \cite{trainingdata_pro_mask_detection} & 300,988 & 75,247 & Private & - \\
    MFI \cite{ding2020masked} & 4,916 & 669 & Private & 2020 \\
    MFV \cite{ding2020masked} & 400 & 200 & Private & 2020\\
    COMASK20 \cite{vu2022masked} & 2754 & 300 & Public & 2022\\
    MDMFR (FMD) \cite{ullah2022novel} & 2896 & 226 & Public & 2022 \\
    MDMFR (MFR) \cite{ullah2022novel} & 6006  & 2 & Public & 2022 \\
    MFDD \cite{10036007} & 24,771 & 2 & Private & 2020 \\
    FMD (AIZOOTech)  \cite{AIZOOTech} & 7,971 & 2 & Public & 2020 \\
    MAFA \cite{ge2017detecting} & 30811 & 3 & Public & 2017 \\
    FMLD \cite{app11052070} & 41,934 & 3 & Public & 2021 \\
    Sunil's custom dataset \cite{singh2021face} & 7,500 & 2 & Public & 2021 \\
    Jun's practical  dataset \cite{zhang2021novel} & 4,672 & 3 & Private & 2021 \\
    ISL-UFMD \cite{eyiokur2022unconstrained} & 21,316 & 3 & Public & 2021 \\
    MD (Kaggle) \cite{make_ml} & 853 & 3 & Public & 2019 \\
    PWMFD \cite{jiang2021real} & 9,205 & 3 & Public & 2021 \\
    Moxa3K \cite{roy2020moxa} & 3,000 & 2 & Public & 2020 \\
    TFM \cite{benitez2022tfm} & 107,598 & 2 & Private & 2022 \\
    WMD \cite{wang2021hybrid} & 7,804 & 1 & Public & 2021 \\
    WMC \cite{wang2021hybrid} & 38,145 & 2 & Public & 2021 \\
    BAFMD \cite{kantarci2022bias} & 6,264 & 2 & Public & 2022 \\
    \hline
\end{tabular}
\end{adjustbox}
\caption{\label{tab:real}
Summary of widely used benchmark datasets featuring real masks.}
\end{table}

In the realm of real face mask datasets, the Real-World Masked Face Recognition Dataset (RMFRD) \cite{Wang2020MaskedFaceRecognition} stands out as one of the most extensive publicly available resources for MFR. RMFRD comprises 90,000 unmasked faces and 5,000 masked faces for 525 individuals, serving as a valuable asset for training and evaluating MFR models. Figure \ref{fig:a} showcases samples from RMFRD, illustrating examples of both masked and unmasked faces for various individuals, thereby facilitating MFR applications and serving as ground truth data for training face mask removal models. Another notable dataset is the Face Mask Detection Dataset \cite{trainingdata_pro_mask_detection}, a large private repository comprising 300,988 images of 75,247 individuals. Each individual in this dataset is represented by four selfie images: one without a mask, one with a properly masked face, and two with incorrectly masked faces. These diverse scenarios offer rich training data for face mask detection models. Additionally, MFR2 \cite{anwar2020masked} is a popular dataset in the field, containing 269 images for 53 identities, typically used to evaluate model performance trained on either real masked datasets or large synthetic datasets. Figure \ref{fig:b} showcases samples from MFR2.

The Masked Face Segmentation and Recognition (MFSR) dataset \cite{geng2020masked} comprises two components. The first part, MFSR-SEG, contains 9,742 images of masked faces sourced from the Internet, each annotated with manual segmentation labels delineating the masked regions. These annotations are particularly useful in FU tasks, serving as an initial step, as seen in stage 1 of GANMasker \cite{mahmoud2023ganmasker}. The second part, MFSR-REC, encompasses 11,615 images representing 1,004 identities, with 704 identities sourced from real-world collections and the remaining images gathered from the Internet. Each identity is represented by at least one image featuring both masked and unmasked faces. F. Ding, P. Peng, et al. \cite{ding2020masked} introduced two datasets to assess MFR models. The first dataset, known as Masked Face Verification (MFV), includes 400 pairs representing 200 distinct identities. The second dataset, Masked Face Identification (MFI), comprises 4,916 images, each corresponding to a unique identity, totaling 669 identities. The COMASK20 dataset \cite{vu2022masked} employed a distinct approach. Video recordings of individuals in various settings and poses were captured and subsequently segmented into frames every 0.5 seconds, each stored in individual folders. Careful manual curation was conducted to remove any obscured images, ensuring data quality. Ultimately, a collection of 2,754 images representing 300 individuals was assembled.

Figure \ref{fig:rmds} presents a selection of datasets specifically designed for Facial Mask Recognition (FMR), each offering samples showcasing correctly masked, incorrectly masked, and unmasked faces. For instance, the Masked Face Detection Dataset (MFDD) \cite{10036007} encompasses 24,771 images of masked faces. This dataset is compiled using two approaches: the first involves integrating data from the AI-Zoo dataset \cite{AIZOOTech}, which itself is a popular FMR dataset comprising 7,971 images sourced from the WIDER Face \cite{yang2016wider} and MAFA \cite{ge2017detecting} datasets. The second approach involves gathering images from the internet. MFDD serves the purpose of determining whether an individual is wearing a mask. The MAFA dataset\cite{ge2017detecting} stands out as a prominent solution to the scarcity of large datasets featuring masked faces. Comprising 30,811 internet-sourced images, each possessing a minimum side length of 80 pixels, MAFA offers a substantial resource for research and development in MFR. The dataset contains a total of 35,806 masked faces, with the authors ensuring the removal of images featuring solely unobstructed faces to maintain focus on occluded facial features.

N. Ullah, A. Javed, et al. \cite{ullah2022novel} introduced a comprehensive dataset named the Mask Detection and Masked Facial Recognition (MDMFR) dataset, aimed at evaluating the performance of both face mask detection and masked facial recognition methods. The dataset comprises two distinct parts, each serving a specific purpose. The first part is dedicated to face mask detection and includes 6,006 images, with 3,174 images featuring individuals wearing masks and 2,832 images featuring individuals without masks. Conversely, the second part focuses on masked facial recognition and contains a total of 2,896 masked images representing 226 individuals. Figure \ref{fig:c} provides visual samples of both categories, with samples of masked and unmasked faces displayed on the left and masked face samples of three different individuals shown on the right.

Researchers continue to leverage existing large datasets like MAFA and Wider Face to develop new datasets tailored to specific research objectives. Batagelj, Peer, et al. \cite{app11052070} introduced the Face-Mask Label Dataset (FMLD), a challenging dataset comprising 41,934 images categorized into three classes: correct mask (29,532 images), incorrect mask (1,528 images), and without mask (32,012 images). Similarly, Singh, S., Ahuja, U., et al. \cite{singh2021face} amalgamated data from MAFA, Wider Face, and additional manually curated images sourced from various online platforms to create a dataset encompassing 7,500 images, referred to here as Sunil's custom dataset. Additionally, J. Zhang, F. Han, et al. \cite{zhang2021novel} developed a practical dataset known here as Jun's practical dataset, comprising a total of 4,672 images. This dataset includes 4,188 images sourced from the public MAFA dataset and 484 images sourced from the internet. The images in Jun's practical dataset are categorized into five types: clean face, hand-masked face, non-hand-masked face, masked incorrect face, and masked correct face, with the first three types grouped into the without-mask class and the remaining two classes designated as mask\_correct and mask\_incorrect. Likewise, the Interactive Systems Labs Unconstrained Face Mask Dataset (ISL-UFMD) \cite{eyiokur2022unconstrained} is compiled from a variety of sources, including publicly available face datasets like Wider-Face \cite{yang2016wider}, FFHQ \cite{karras2019style}, CelebA \cite{liu2015faceattributes}, and LFW \cite{huang2008labeled}, in addition to YouTube videos and online sources. This comprehensive dataset comprises 21,316 face images designed for face-mask detection scenarios, encompassing 10,618 images of individuals wearing masks and 10,698 images of individuals without masks.

Moreover, several other datasets cater to face mask detection tasks. For instance, the Masked dataset (MD-Kaggle) \cite{make_ml} hosted on Kaggle comprises 853 images annotated into three categories. Similarly, the Properly Wearing Masked Face Detection Dataset (PWMFD) \cite{jiang2021real} contains 9,205 images categorized into three groups. Additionally, datasets like Moxa3K \cite{roy2020moxa} and TFM \cite{benitez2022tfm} focus on binary classifications, distinguishing between images with and without masks. Moxa3K features 3,000 images, while TFM boasts a larger private collection of 107,598 images. Furthermore, the Wearing Mask Detection (WMD) \cite{wang2021hybrid} dataset provides 7,804 images for training detection models. Meanwhile, the Wearing Mask Classification (WMC) \cite{wang2021hybrid} dataset consists of two classes, with 19,590 images containing masked faces and 18,555 background samples, resulting in a total of 38,145 images. The Bias-Aware Face Mask Detection (BAFMD) dataset \cite{kantarci2022bias} comprises 6,264 images, featuring 13,492 faces with masks and 3,118 faces without masks. Noteworthy is that each image contains multiple faces.

\begin{figure*}[h] 
  \centering
  \includegraphics[width=\textwidth,height=.5\textheight]{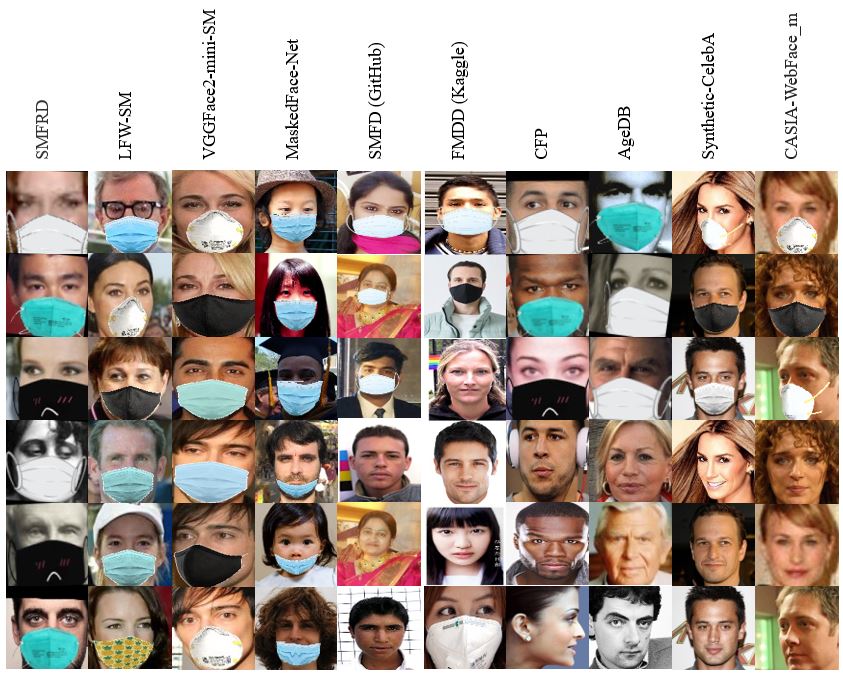}
  \caption{Samples of synthetic masked faces from benchmark datasets.}
  \label{fig:synthetic-sampels}
\end{figure*}

\subsection{Synthetic Mask Datasets}
However, the accessibility of synthetic masked face datasets has seen a significant boost due to the abundance of public face datasets available for generation purposes. In contrast to datasets featuring authentic masks, synthetic mask datasets bring a unique perspective by incorporating artificially generated face coverings. These datasets offer a controlled environment, enabling researchers to explore various synthetic mask variations, including considerations such as style, color, and shape. The controlled nature of these datasets facilitates a systematic exploration of challenges in MFR, providing valuable insights into the response of models to different synthetic masks. Within this subsection, we delve into benchmark datasets featuring synthetic masks, examining their creation processes, advantages, and potential applications in the field of masked face recognition and related tasks. Figure \ref{fig:synthetic-sampels} showcases samples from some datasets with synthetic masks, while Table \ref{tab:freq} provides a detailed summary of these synthetic datasets discussed in this subsection.

As highlighted in the preceding subsection, RMFRED stands out as one of the most expansive real-mask masked face datasets. Z. Wang, B. Huang, et al \cite{10036007} took a different approach by automatically applying masks to face images sourced from existing public datasets such as CASIA-WebFace \cite{yi2014learning}, LFW \cite{huang2008labeled}, CFP-FP \cite{sengupta2016frontal}, and AgeDB-30 \cite{moschoglou2017agedb}. This effort resulted in creating a Simulated Mask Face Recognition Dataset (SMFRD), comprising 536,721 masked faces representing 16,817 unique identities. Aqeel Anwar and Arijit Raychowdhury \cite{anwar2020masked} introduced MaskTheFace, an open-source tool designed to mask faces in public face datasets effectively. This tool was utilized to generate large masked face datasets like LFW-SM, which contains 64,973 images representing 5,749 identities. It incorporates various types of masks, including surgical-green, surgical-blue, N95, and cloth, derived from the original LFW dataset \cite{huang2008labeled}. Utilizing the same tool, they developed another masked dataset called VGGFace2-mini-SM \cite{anwar2020masked}, extracted from the original VGGFace2-mini dataset, a subset of the VGGFace2 dataset \cite{cao2018vggface2}, with 42 images randomly selected per identity. This augmentation expanded the total image count to 697,084, maintaining the same 8,631 identities. Various mask types, akin to those employed in LFW-SM, were integrated into this dataset.

\begin{table}[htbp]
    \centering
  \begin{adjustbox}{width={0.5\textwidth}}
  \begin{tabular}{c|c|c|c|c}
    \hline
    \textbf{Dataset} & \textbf{Size} & \textbf{Identities} & \textbf{Access} & \textbf{Year} \\
    \hline
    SMFRD \cite{10036007} & 536,721 & 16,817 & Public & 2020 \\
    \hline
    LFW-SM \cite{anwar2020masked} & 64,973 &  5,749 & Public & 2020 \\
    VGGFace2-mini-SM \cite{anwar2020masked} & 697,084 & 8,631 & Public & 2020 \\
    MaskedFace-Net \cite{cabani2021maskedface} & 137,016 & 2 & Public & 2022 \\
    SMFD(GitHub) \cite{SMFD-Github} & 1,376 & 2 & Public & 2020 \\
    FMDD(Kaggle) \cite{FMDD-kaggle} & 7,553 & 2 & Public & 2020 \\
    MS1MV2-Masked \cite{boutros2022self} & 5.374 M & 85,000 & Public & 2022 \\
    LFW \cite{huang2008labeled} & 13,233 & 5,749 & Public & 2008 \\
    IJB-C \cite{maze2018iarpa} & 148,876 & 3531 & Public & 2018 \\
    CFP \cite{sengupta2016frontal} & 7,000 & 500 & Public & 2016 \\
    AgeDB \cite{moschoglou2017agedb} & 16, 488 & 568 & Public & 2017 \\
    CelebA \cite{liu2015faceattributes} & +200,000 & 10,000 & Public & 2015 \\
    CelebA-HQ \cite{karras2017progressive} & 30,000 & - & Public & 2018 \\
    Synthetic-CelebA \cite{mahmoud2023ganmasker} & 30,000 & - & Private & 2023 \\
    PS-CelebA \cite{din2020novel} & 10,000 & - & Private & 2020 \\    
    VGGFace2 \cite{cao2018vggface2} & 3.31M & 9,131 & Public & 2018 \\
    VGGFace2\_m \cite{deng2021mfcosface} & 666,800 & 8,335 & Public & 2021 \\
    LFW\_m \cite{deng2021mfcosface} & 26,466 & 8,5749 & Public & 2021 \\
    CF\_m \cite{deng2021mfcosface} & 5,000 & 500 & Public & 2021 \\
    CASIA-WebFace \cite{yi2014learning} & 494,414 & 10,575 & Public & 2014 \\
    CASIA-WebFace\_m \cite{pann2022effective} & 789,296 & 10,575 & Public & 2022 \\
    \hline
\end{tabular}
\end{adjustbox}
\caption{\label{tab:freq} summary of popular Synthetic mask benchmarking datasets.}
\end{table}

F. Boutros, N. Damer, et al. \cite{boutros2022self} developed a synthetic dataset named MS1MV2-Masked, derived from MS1MV2 \cite{deng2019arcface}, incorporating various mask shapes and colors. They utilized Dlib \cite{king2009dlib} for extracting facial landmark points. However, they noted that Dlib encountered difficulties in extracting landmarks from 426 images. Consequently, their synthetic dataset comprises approximately 5,374 images representing 85,000 identities. 

Building upon the FFHQ dataset \cite{karras2019style}, C. Adnane, H. Karim, et al. \cite{cabani2021maskedface} introduced a comprehensive synthetic dataset called MaskedFace-Net. This dataset comprises two primary sub-datasets: the Correctly Masked Face Dataset (CMFD) and the Incorrectly Masked Face Dataset (IMFD). These datasets serve as the sole categories within the main dataset dedicated to face mask detection, collectively containing a total of 137,016 images. Notably, the dataset is comprised of 49\% correctly masked faces (67,193 images) and 51\% incorrectly masked faces (69,823 images). The SMFD dataset \cite{SMFD-Github} comprises two distinct classes: one with a mask, encompassing 690 images, and the other without a mask, comprising 686 images. Consequently, the dataset contains a total of 1,376 images, which were utilized to train a classification CNN network tasked with distinguishing between faces with masks and those without. Moreover, the Face Mask Detection Dataset \cite{FMDD-kaggle} comprises 7,553 images divided into two categories: with masks and without masks. Specifically, there are 3,725 images depicting faces with masks and 3,828 images featuring faces without masks. These images were sourced from the internet and encompass the entirety of images found in the SMFD dataset \cite{SMFD-Github}. Thus, the dataset represents a hybrid composition, incorporating both real masked faces and synthetic masked faces. They employed this dataset to train their model and also utilized a similar approach to create additional synthetic datasets based on Labeled Faces in the Wild (LFW) \cite{huang2008labeled} and IARPA Janus Benchmark -C (IJB-C) \cite{maze2018iarpa}.

It is common practice among authors to create synthetic datasets either manually or automatically using deep learning tools. Therefore, various face datasets can be considered or referenced in synthetic masked face datasets. One such dataset is the Celebrities in Frontal-Profile (CFP) dataset \cite{sengupta2016frontal}, comprising 7,000 images representing 500 identities. This dataset is divided into two sub-datasets based on the face angle: frontal faces and profile faces, both depicting the same 500 identities. The frontal faces subset consists of 10 images per identity, while each identity in the profile faces subset is represented by 4 images. The AgeDB \cite{moschoglou2017agedb} dataset comprises 16,488 images featuring 568 prominent individuals, including actors/actresses, politicians, scientists, writers, and more. Each image comes annotated with identity, age, and gender attributes. Moreover, CelebA \cite{liu2015faceattributes} stands out as one of the largest face datasets, boasting over 200,000 images spanning 10,000 identities. Complementing this dataset is CelebA-HQ \cite{karras2017progressive}, a high-resolution variant derived from CelebA, featuring 30,000 meticulously crafted high-quality images. Furthermore, leveraging the CelebA dataset, M. Mohamed and K. Hyun-Soo \cite{mahmoud2023ganmasker} curated a Synthetic Masked Dataset comprising 30,000 images distributed across three subfolders: original unmasked faces, masked faces, and binary mask maps. Employing the MaskTheFace tool \cite{din2020novel}, they generated two masked datasets with varying sizes (256 and 512 pixels) tailored for face mask removal tasks. Additionally, N. Ud Din, K. Javed, et al. [4] engineered a synthetic dataset using Adobe Photoshop CC 2018, derived from CelebA, featuring 10,000 masked images alongside their original counterparts.

Deng H, Feng Z, et al. \cite{deng2021mfcosface} introduced three composite datasets featuring mixed masked and unmasked faces. These datasets integrate original images with masked faces generated through their proprietary masked-face image generation algorithm. The first dataset, VGGFace2\_m \cite{deng2021mfcosface}, originates from the VGGFace2 \cite{cao2018vggface2} face dataset, leveraging 8,335 identities from the VGG-Face2 training set. From each identity, 40 pictures were randomly selected to construct VGGFace2\_mini, which was then combined with the generated masked faces to form VGGFace2\_m. Similarly, LFW\_m \cite{deng2021mfcosface} was crafted using the LFW dataset \cite{huang2008labeled}, a widely used benchmark for facial recognition, comprising 13,233 face images and 5,749 identities. The masked-face images generated were merged with the original LFW dataset to produce LFW\_m. Lastly, CF\_m \cite{deng2021mfcosface} was derived from the CASIA-FaceV5 dataset \cite{xiong2018asian}, which features images of 500 individuals, with five images per person totaling 2,500 images. The original images from CASIA-FaceV5 were amalgamated with masked images generated by their algorithm to create CF\_m. Furthermore, Pann V and Lee HJ \cite{pann2022effective} devised CASIA-WebFace\_m as an extension of the CASIA-WebFace dataset \cite{yi2014learning}, a comprehensive public face recognition dataset encompassing 494,414 images representing 10,575 distinct identities. However, due to limitations with the data augmentation tool, their generated masked faces amounted to 394,648, with 20\% of face images remaining undetected. These generated masked face images were then integrated with the corresponding unmasked images from the original dataset, resulting in CASIA-WebFace\_m for model training purposes. Consequently, the combined dataset boasts a total of 789,296 training samples. Moreover, they produced modified versions of the original datasets LFW \cite{huang2008labeled}, AgeDB \cite{moschoglou2017agedb}, and CFP \cite{sengupta2016frontal}, labeled as LFW\_m, AgeDB-30\_m, and CFP-FP\_m, respectively.

\section{Evaluation Metrics}
This section will explain standard evaluation metrics, focusing specifically on those applied in MFR, FMR, and Face Unmasking (Face Mask Removal). Evaluating model performance in these domains is essential for gauging their effectiveness in real-world applications. To this end, various evaluation metrics and benchmarking strategies are utilized to assess accuracy, robustness, and efficiency. In the following discussion, we explore the primary evaluation metrics and benchmarking approaches employed in these tasks.

\begin{enumerate}

    \item \textbf{Accuracy} is a fundamental evaluation metric utilized across various domains, including facial recognition tasks. It represents the proportion of correct predictions relative to the total number of samples and can be formally defined as illustrated in Equation \ref{eq:Accuracy}.
        \begin{equation} \label{eq:Accuracy}
           Accuracy = \frac{TP + TN}{TP + TN + FP + FN}
        \end{equation}
    
    \item \textbf{ERR (Error rate)} is a crucial metric utilized in diverse classification tasks, offering valuable insights into model accuracy by measuring misclassifications relative to dataset size. Unlike accuracy, ERR accounts for both false positives and false negatives, providing a comprehensive assessment of model performance. Its sensitivity to imbalanced data underscores its importance, making it an essential tool for evaluating classification accuracy. Mathematically, ERR is calculated by dividing the sum of false positive and false negative predictions by the total number of instances, as shown in Equation \ref{eq:ERR}.
        \begin{equation} \label{eq:ERR}
            \small
           ERR = \frac{FP + FN}{TP + TN + FP + FN} = 1 - Accuracy
        \end{equation}
        
    \item \textbf{Precision} quantifies the proportion of accurate positive identifications among all the positive matches detected, and it can be formally expressed as depicted in Equation \ref{eq:Precision}.
        \begin{equation} \label{eq:Precision}
           Precision = \frac{TP}{TP + FP}
        \end{equation}
      
    \item \textbf{Recall} also known as \textbf{sensitivity} or \textbf{True Positive Rate} measures the proportion of true positive instances correctly identified by the system out of all actual positive instances. It is formally defined as shown in Equation \ref{eq:Recall}.
        \begin{equation} \label{eq:Recall}
           Recall = \frac{TP}{TP + FN}
        \end{equation}

    \item \textbf{F1-Score} is a pivotal evaluation metric, that represents the harmonic mean of precision and recall. This metric offers a balanced measure of the facial recognition model's performance, accounting for both false positives and false negatives. Particularly valuable for imbalanced datasets, the F1-Score provides a comprehensive assessment of model performance. Unlike accuracy, which may overlook certain types of errors, the F1-Score considers both false positives and false negatives, rendering it a more reliable indicator of a model’s effectiveness. Its calculation is demonstrated in Equation \ref{eq:F1}.
        \begin{equation} \label{eq:F1}
           F1-Score = 2 \times \frac{Precision \times Recall}{Precision + Recall}
        \end{equation}

    \item \textbf{ROC (Receiver Operating Characteristic) Curve} ROC curves \cite{zweig1993receiver} graphically represent the trade-off between sensitivity (true positive rate) and specificity (true negative rate) across various threshold values. This visualization aids in selecting an optimal threshold that strikes a balance between true positive and false positive recognition rates. By examining the ROC curve, decision-makers can effectively assess the performance of a classification model and make informed decisions about threshold selection.

    \item \textbf{AUC (Area Under the Curve)} is a pivotal evaluation metric in classification tasks, offering a comprehensive assessment of a model's performance. It quantifies the discriminative power of the model across varying threshold values, providing insights into its ability to correctly classify positive and negative instances. A higher AUC value signifies stronger discrimination, indicating superior model performance. Conversely, an AUC value of 0.5 suggests that the model's predictive ability is no better than random chance. AUC is instrumental in gauging the effectiveness of classification models and is widely utilized in performance evaluation across diverse domains.

    \item \textbf{Confusion Matrix} provides a detailed breakdown of the model's predictions, including true positives, true negatives, false positives, and false negatives. It serves as a basis for computing various evaluation metrics and identifying areas for improvement.

    \item \textbf{FAR (False Acceptance Rate)} serves as a focused gauge of security vulnerabilities, offering precise insights into the system's efficacy in thwarting unauthorized access attempts. This pivotal metric plays a crucial role in evaluating the overall security effectiveness of biometric authentication systems, thereby guiding strategic endeavors aimed at bolstering system reliability and mitigating security threats. Equation \ref{eq:FAR} delineates its formula, providing a quantifiable framework for assessing system performance.
        \begin{equation} \label{eq:FAR}
           FAR = \frac{FP}{FP+TN}
        \end{equation}

    \item \textbf{FRR (False Rejection Rate)} is a crucial metric for evaluating system usability, representing the likelihood of the system inaccurately rejecting a legitimate identity match. Its assessment is integral to gauging the user-friendliness of the system, with a high FRR indicating diminished usability due to frequent denial of access to authorized individuals. Conversely, achieving a lower FRR is essential for improving user satisfaction and optimizing access procedures. The calculation of FRR is depicted by Equation \ref{eq:FRR}.
        \begin{equation} \label{eq:FRR}
           FRR = \frac{FN}{FN+TP}
        \end{equation}    

    \item \textbf{EER (Equal Error Rate)}  denotes the threshold on the ROC curve where the False Acceptance Rate (FAR) equals the False Rejection Rate (FRR), signifying the equilibrium point between false acceptance and false rejection rates. A lower EER signifies superior performance in achieving a balance between these two error rates.

    \item \textbf{Specificity} also known as the \textbf{True Negative Rate} gauges the system's proficiency in accurately recognising negative instances. Specifically, it assesses the system's capability to correctly identify individuals who are not the intended subjects. Mathematically, specificity is calculated using Equation \ref{eq:Specificity}. This metric offers valuable insights into the system's performance in correctly classifying negatives, contributing to its overall effectiveness and reliability.
        \begin{equation} \label{eq:Specificity}
           Specificity = \frac{TN}{TN + FP} = 1 - FAR
        \end{equation}

    \item \textbf{Rank-N Accuracy} is a widely employed metric in facial recognition tasks, that assesses the system's capability to prioritize the correct match within the top-N retrieved results. It quantifies the percentage of queries for which the correct match is positioned within the top-N-ranked candidates. In the Rank-N Identification Rate evaluation, the system's output is considered accurate if the true identity of the input is within the top N identities listed by the system. For instance, in a Rank-1 assessment, the system is deemed correct if the true identity occupies the top spot. Conversely, in a Rank-5 evaluation, the system is considered accurate if the true identity is among the top 5 matches. A higher Rank-N Accuracy signifies superior performance in identifying the correct match among the retrieved candidates, providing valuable insights into the system's efficacy in real-world scenarios. Mathematically, it is represented as depicted in Equation \ref{eq:Rank-N}.
    
        \begin{equation} \label{eq:Rank-N}
            \small
            \text{Rank-N Accuracy} = \frac{\text{\# correct matches within top-N}}{\text{Total number of queries}}
        \end{equation}

    \item \textbf{Intersection over Union (IoU)} quantifies the extent of spatial overlap between the predicted bounding box (P) and the ground truth bounding box (G). Its mathematical representation is shown in Equation \ref{eq:IoU}.
        \begin{equation} \label{eq:IoU}
           IoU = \frac{P\cap G}{P\cup G}
        \end{equation}

    \item \textbf{AP (Average Precision)} serves as a crucial measure in assessing object detection systems. It provides insight into how effectively these systems perform across different confidence thresholds, by evaluating their precision-recall performance. AP computes the average precision across all recall values, indicating the model's ability to accurately detect objects at varying confidence levels. This calculation involves integrating the precision-recall curve, as demonstrated in Equation \ref{eq:AP}. By considering the precision-recall trade-off comprehensively, AP offers a holistic evaluation of the detection method's effectiveness.
        \begin{equation} \label{eq:AP}
           AP = \int_{0}^{1} P(r) dr
        \end{equation}
    Where $ P(r)$ represents the precision at a given recall threshold r, where r ranges from 0 to 1.

    \item \textbf{mAP (Mean Average Precision)} enhances the notion of AP by aggregating the average precision values across multiple object classes. It offers a unified metric summarizing the overall performance of the object detection model across diverse object categories. Mathematically, mAP is calculated as the average of AP values for all classes, as illustrated in Equation \ref{eq:mAP}.
        \begin{equation} \label{eq:mAP}
           mAP = \frac{1}{N} \sum_{i=1}^{N} AP_i
        \end{equation}

    \item \textbf{Dice Score} also known as \textbf{Dice Coefficient} is a metric commonly used in image segmentation tasks to assess the similarity between two binary masks or segmentation maps. It quantifies the spatial overlap between the ground truth mask (A) and the predicted mask (B), providing a measure of segmentation accuracy. The Dice Score equation compares the intersection of  A and B with their respective areas, as defined in Equation \ref{eq:Dice-Score}.
        \begin{equation} \label{eq:Dice-Score}
           Dice\ Score = \frac{2\times(|A\cap B|)}{|A|+|B|}
        \end{equation}
    Where $|A\cap B|$ represents the number of overlapping pixels between the ground truth and predicted masks, while $|A|$ and $|B|$ denote the total number of pixels in each mask, respectively.
        
    \item \textbf{PSNR (Peak Signal-to-Noise Ratio)} is widely employed in image inpainting to assess the quality of image generation or reconstruction. It quantifies the level of noise or distortion by comparing the maximum possible pixel value to the mean squared error (MSE) between the original and reconstructed images, as depicted in Equation 13.
        \begin{equation} \label{eq:PSNR}
           PSNR = 10\log_{10}(\frac{MAX^2}{MSE})
        \end{equation}

    \item \textbf{SSIM (Structural Similarity Index Measure)} The structural Similarity Index Measure \cite{wang2004image} evaluates the similarity between two images by considering their luminance, contrast, and structure. It provides a measure of perceptual similarity, accounting for both global and local image features. SSIM is calculated by comparing the luminance, contrast, and structure similarity indexes, as expressed in Equation \ref{eq:SSIM}.
        \begin{equation} \label{eq:SSIM}
           SSIM(x,y) = \frac{(2\mu_{x}\mu_{y} + C_1) (2 \sigma _{xy} + C_2)} 
    {(\mu_{x}^2 + \mu_{y}^2+C_1) (\sigma_{x}^2 + \sigma_{y}^2+C_2)}
        \end{equation}
    Where x and y represent the two compared images, $\mu_{x}$  and $\mu_{y}$  denote the mean of x and y respectively, $\sigma_{x}^2$  and $\sigma_{y}^2$  represent the variances of x and y respectively, $\sigma_{xy}$ is the covariance of x and y, and $C_1$  and $C_2$  are constants to stabilize the division, typically set to small positive values.   
    
    \item \textbf{FID (Fréchet Inception Distance)} \cite{heusel2017gans} serves as a metric for assessing the likeness between two sets of images. It quantifies the disparity between feature representations of real and generated images within a learned high-dimensional space, typically modeled by a pre-trained neural network. A lower FID score denotes a higher degree of resemblance between the datasets. The calculation of FID involves the application of the Fréchet distance formula, as depicted in Equation \ref{eq:FID}.

        \begin{equation} \label{eq:FID}
            \small
           FID = ||\mu_{R} - \mu_{F}||^2 + Tr(C_{R} + C_{F} - 2(C_{R} - C_{F})^\frac{1}{2})
        \end{equation}
    where $\mu_{R}$  and $\mu_{F}$  are the mean feature vectors of the real and generated image sets, $C_{R}$  and $C_{F}$ are their covariance matrices, and Tr denotes the trace operator.
    
    \item \textbf{NIQE (Naturalness Image Quality Evaluator)} \cite{mittal2012making} assesses the quality of an image based on natural scene statistics. It evaluates the level of distortions introduced during image acquisition or processing, providing a measure of image fidelity. NIQE computes the deviation of the image from the expected natural scene statistics, with higher scores indicating greater image distortion.
        
    \item \textbf{BRISQUE (Blind/Referenceless Image Spatial Quality Evaluator)} \cite{mittal2012no} is a no-reference image quality assessment metric. It evaluates the perceived quality of an image by analyzing its spatial domain features, such as local sharpness and contrast. BRISQUE computes a quality score based on the statistical properties of these features, with lower scores indicating higher image quality.
    
\end{enumerate}

\section{ Masked Faces Methods}

With the increased use of face masks in response to the COVID-19 outbreak, researchers have focused on the challenges given by masked faces, particularly in the domains of MFR, FMR, and FU. This section thoroughly analyzes the most recent breakthroughs in deep learning-based state-of-the-art (SOTA) approaches to overcoming these challenges. With three distinct subsections dedicated to each task, or preliminary steps toward them, ranging from face mask detection and removal to masked face recognition approaches, researchers have introduced a diverse array of innovative solutions to improve the accuracy and reliability of masked face recognition systems. By thoroughly exploring these designs and approaches, this section aims to provide important insights into current advancements in deep learning-based approaches for masked face-related tasks and elucidate potential avenues for future research in these rapidly evolving fields.

\begin{figure*}[htbp]
  \centering
  \includegraphics[width=\linewidth]{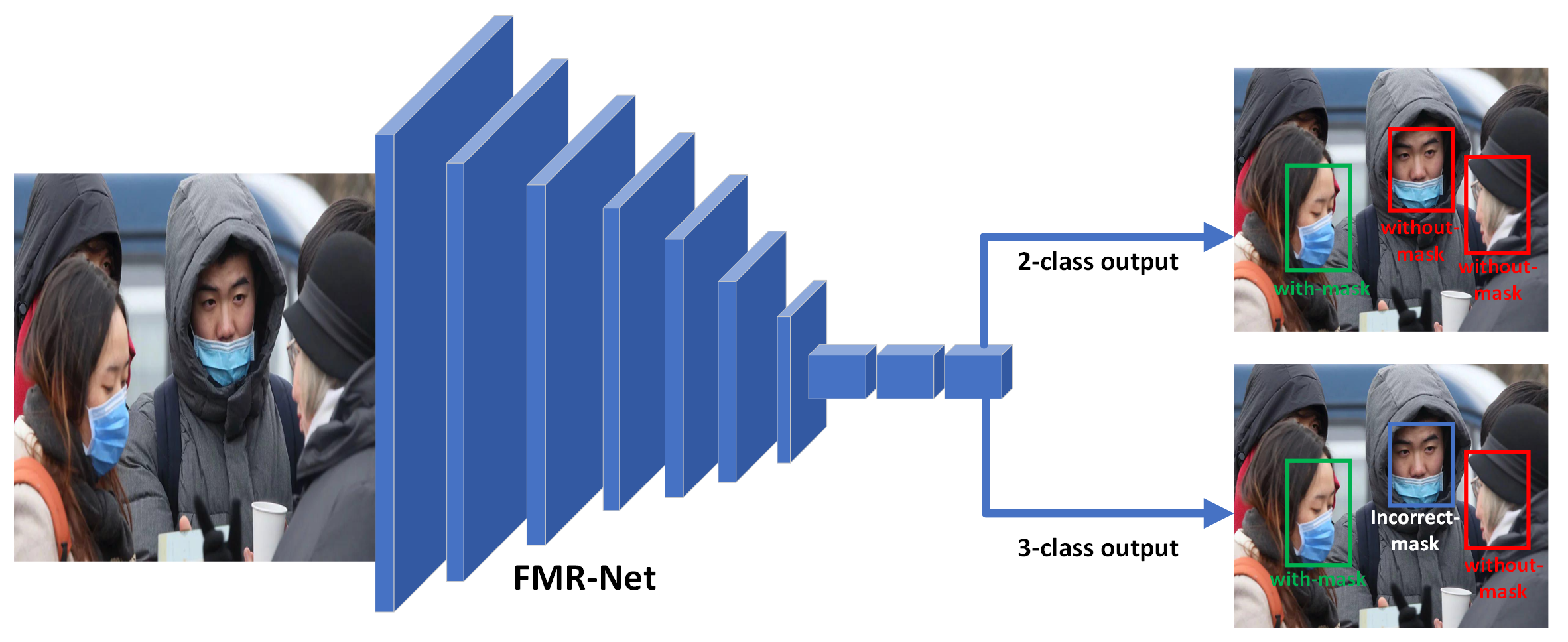}
  \caption{Illustration of the FMR-Net Architecture for Face Mask Recognition, Depicting Two-Subtask Scenarios: 2-Class (With and Without Mask) and 3-Class (With, Incorrect, and Without Mask).}
  \label{fig:3}
\end{figure*}

\subsection{Face Mask Recognition Approaches}
In the realm of computer vision, face mask detection has become crucial, especially during health crises like the COVID-19 pandemic. This technology relies on machine learning (ML) and deep learning (DL) techniques to automatically detect masks on human faces. DL, particularly Convolutional Neural Networks (CNNs), excels by directly extracting features from raw input data, eliminating the need for manual feature engineering. Various backbone architectures, including Multi-stage, YOLO-based, and Transfer Learning, hierarchically process data to distinguish masked from unmasked faces. The choice of backbone architecture significantly impacts accuracy and computational efficiency\cite{habib2022efficient,boulila2021deep}.

Table \ref{tab:Face-Mask-Recognition-models},\ref{tab:Face-Mask-Recognition-models2} summarizes various face mask recognition models evaluated using different datasets and performance metrics. Furthermore, Figure \ref{fig:3} showcases the FMR-Net architecture, depicting examples of both the two-subtask 2-class scenario, distinguishing between with-mask and without-mask and the 3-class scenario, discerning between with-mask, incorrect-mask, and without-mask.

\subsubsection{Convolutional Neural Network} are crucial in computer vision due to their efficient pattern recognition and spatial feature extraction capabilities. By applying convolutional filters directly to input images, CNNs efficiently isolate high-level features, enhancing both accuracy and computational speed for tasks like image classification and object detection.  FMJM Shamrat et al. \cite{shamrat2021face} exploring three deep-learning techniques for face mask detection: Max pooling, Average pooling, and MobileNetV2. MobileNetV2 achieved the highest accuracies 99.72\% in training and 99.82\% in validation—demonstrating robust capability. While H Goyal et al.\cite{goyal2022real} developed an automated face mask detection model to enforce mask-wearing in public spaces. This model, capable of processing both static images and real-time video feeds, classifies subjects as ‘with mask’ or ‘without mask.’ Trained and evaluated using a dataset of approximately 4,000 images from Kaggle, the model achieved an accuracy rate of 98\%. It demonstrated computational efficiency and precision.

\subsubsection{ Multi-stage detection} is a category of object detection algorithms where the detection process is divided into several sequential steps. In a typical multi-stage detector, such as RCNN, the first step involves identifying a set of potential regions of interest within an image, often through a technique like selective search. Subsequently, each region is individually processed to extract CNN feature vectors, which are then used to classify the presence and type of objects within those regions. This approach contrasts with single-stage detectors that perform detection in a single pass without a separate region proposal phase, trading some accuracy for increased processing speed.

S Sethi et al.\cite{sethi2021face} introduces a novel real-time face mask detection technique. Combining one-stage and two-stage detectors, it accurately identifies individuals not wearing masks in public settings, supporting mask mandate enforcement. Utilizing ResNet50 as a baseline, transfer learning enhances feature integration across levels, and a new bounding box transformation improves localization accuracy. Experiments with ResNet50, AlexNet, and MobileNet optimize the model’s performance, achieving 98.2\% accuracy. 
    
Also A Chavda et al. \cite{chavda2021multi} develop two-stage architecture combines RetinaFace face detection with a CNN classifier. Faces detected by RetinaFace are processed to determine mask presence. The classifier, trained using MobileNetV2, DenseNet121, and NASNet, ensures efficient real-time performance in CCTV systems. M Umer et al.\cite{umer2023face} developed a new dataset called RILFD, consisting of real images annotated with labels indicating mask usage. Unlike simulated datasets, RILFD provides a more accurate representation for training face mask detection systems. The researchers evaluated machine learning models, including YOLOv3 and Faster R-CNN, adapting them specifically for detecting mask-wearing individuals in surveillance footage. Enhancing these models with a custom CNN and a four-step image processing technique, they achieved an impressive 97.5\% accuracy on the RILFD dataset, as well as on two other publicly available datasets (MAFA and MOXA). 
    
\subsubsection{Single Shot Detector} is an object detection technique that streamlines the process by using a single deep neural network. Unlike methods that rely on a separate region proposal network (which can slow down processing), SSD directly predicts object bounding boxes and class labels in a single pass. This efficiency allows SSD to process images in real time with high speed and accuracy. S Vignesh Baalaji et al.\cite{vignesh2023autonomous} proposes an autonomous system for real-time face mask detection during the COVID-19 pandemic. Leveraging a pre-trained ResNet-50 model, the system fine-tunes a new classification layer to distinguish masked from non-masked individuals. Using adaptive optimization techniques, data augmentation, and dropout regularization, the system achieves high accuracy. It employs a Caffe face detector based on SSD to identify face regions in video frames. Faces without masks undergo further analysis using a deep siamese neural network (based on VGG-Face) for identity retrieval. The classifier and identity model achieve impressive accuracies of 99.74\% and 98.24\%, respectively.
    
B Sheikh et al.\cite{sheikh2023rrfmds} presents the Rapid Real-Time Face Mask Detection System (RRFMDS) which is an automated method designed to monitor face mask compliance using video surveillance. It utilizes a Single-Shot Multi-Box Detector for face detection and a fine-tuned MobileNetV2 for classifying faces as masked, unmasked, or incorrectly masked. Seamlessly integrating with existing CCTV infrastructure, the RRFMDS is efficient and resource-light, ideal for real-time applications. Trained on a custom dataset of 14,535 images, it achieves high accuracy (99.15\% on training and 97.81\% on testing) while processing frames in just over 0.14 seconds. While P Nagrath et al. \cite{nagrath2021ssdmnv2} developed a resource-efficient face mask detection model using a combination of deep learning technologies including TensorFlow, Keras, and OpenCV. Their model, SSDMNV2, employs a Single Shot Multibox Detector (SSD) with a ResNet-10 backbone for real-time face detection and uses the lightweight MobileNetV2 architecture for classifying whether individuals are wearing masks. They curated a balanced dataset from various sources, enhanced it through preprocessing and data augmentation techniques, and achieved high accuracy and F1 scores.

\begin{table*}[htbp]
  \centering
  \begin{adjustbox}{width={\textwidth}}
  \begin{tabular}{|c|c|c|c|c|c|c|} 
    \hline
    \textbf{Model} & \textbf{Year} & \textbf{Dataset} & \textbf{Accuracy} & \textbf{Precision}  & \textbf{Recall} & \textbf{F1-score} \\
    \hline
    Max pooling \cite{shamrat2021face} & \multirow{3}{*}{2021} & \multirow{3}{*}{RMFD + SMFD + Own Dataset} & 98.67\% &-&-&-\\
     Average pooling &  &   &  96.23\%&-&-&-\\
     MobileNetV2 &  &  &   99.82\%&-&-&-\\
    \hline
    CNN \cite{goyal2022real} & 2021 & Kaggle data-set &  98\% &  98\% &  97\%  &  98\%\\
    \hline
    ResNet50 + bounding box transformation  \cite{sethi2021face} & 2021 & MAFA dataset Face Detection & - &  99.2\% &  99\% & -\\
    \cline{3-7}
    &  & MAFA dataset Mask Detection & -  &  98.92\% & 98.24\% & -\\
    \hline
    RetinaFace + CNN(NASNetMobile)) \cite{chavda2021multi} & \multirow{2}{*}{2021} & \multirow{2}{*}{RMFRD + Larxel (Kaggle)} & 99.23\%\ & 98.28\% & 100\% & 99.13\%\\
    \cline{4-7}
    RetinaFace + CNN(Dense Net121)) &  &    &  99.49\% & 99.70\% & 99.12\% & 99.40\%\\
    \hline
    \multirow{4}{*}{\makecell{Customized CNN + \\Image Preprocessing Techniques \cite{umer2023face}}} &  \multirow{4}{*}{2023} & RILFD &  97.25\% &  96.20\% &  97.34\% &  96.77\%\\
    \cline{3-7}
    &  & MAFA & 95.74 & - &  94.29\%  & -\\
    \cline{3-7}
    &  & MOXA &  94.37\% & -  &  95.28\% & - \\
    \cline{3-7}
    &  & RMFRD & 99.63\% & -  &  99.69\% & - \\    
    \hline
    \makecell{SSD, ResNet-50, and \\DeepSiamese Neural Network \cite{vignesh2023autonomous}} & 2023 & RMFD + Larxel & 98.24\% & -  & -  & - \\
    \hline
    MobileNetV2 and Caffe-based SSD \cite{sheikh2023rrfmds} & 2023 & Efficient Face Mask Dataset & 97.81\%  & -  & -  & 98\% \\
    \hline
    SSDMNV2 \cite{nagrath2021ssdmnv2} & 2021 & Self-made Dataset of Masked Faces & 92.64\%& -  & -& 93\% \\
    \hline
    Fine-Tuning of InceptionV3 \cite{jignesh2020face} & 2020 & SMFD &  100\% &  100\% & -  & -  \\
    \hline
    MobileNetV2 + SVM\cite{oumina2020control} & 2020 & Private Dataset & 97.11\% &  95.08\% &  94.84\% & -  \\
    \hline
    \multirow{3}{*}{Resnet50 + SVM + ensemble algorithm\cite{loey2021hybrid}} & \multirow{3}{*}{2021} & RMFD & 99.64\% & -  & -  & -\\
    \cline{3-7}
    &  & SMFD &  99.49\% & -  & -  & -\\
    \cline{3-7}
    &  & LFW  &  100\% & -  & -  & -\\
    \hline
    \multirow{2}{*}{Faster\_RCNN + InceptionV2 + BLS \cite{wang2021hybrid}} & \multirow{2}{*}{2021} & WMD Simple Scene & - & 96.46\% &  98.20\% &  97.32\%\\
    \cline{3-7}
    &  & WMD Complex Scene & - & 94.22\% &  88.24\% & 91.13\%\\
    \hline
    Fusion Transfer Learning \cite{su2022face} & 2022 & RMFD and MAFA & 97.84\% & - &  97.87\% &   98.13\%\\
    \hline   
    CMNV2\cite{kumar2023face} & 2023 & The Prajna Bhandary & 99.64\% &  100\% & 99.28\% &   99.64\%\\ 
    \hline
  \end{tabular}
  \end{adjustbox}
\caption{\label{tab:Face-Mask-Recognition-models} Summary of different Face Mask Recognition Models}
\end{table*}

\subsubsection{Transfer Learning} is a technique in deep learning where a model trained on one task is repurposed as the starting point for a model on a different but related task. This approach leverages pre-trained networks, such as InceptionV3, to improve learning efficiency and model performance, particularly when data is limited. 

G Jignesh Chowdary et al.\cite{jignesh2020face} propose an automated method for detecting individuals not wearing masks in public and crowded areas during the COVID-19 health crisis. They employ transfer learning with the pre-trained InceptionV3 model, fine-tuning it specifically for this task. Training is conducted on the Simulated Masked Face Dataset (SMFD), augmented with techniques like shearing, contrasting, flipping, and blurring. while A Oumina et al.\cite{oumina2020control} introduced a novel method for detecting whether individuals are wearing face masks using artificial intelligence technologies. They utilized deep Convolutional Neural Networks (CNNs) to extract features from facial images, which were then classified using machine learning algorithms such as Support Vector Machine (SVM) and K-Nearest Neighbors (K-NN). Despite the limited dataset of 1376 images, the combination of SVM with the MobileNetV2 model achieved a high classification accuracy of 97.1\%.

M Loey et al.\cite{loey2021hybrid} propose a hybrid model combining deep learning and classical machine learning techniques to detect face masks, a vital task during the COVID-19 pandemic. The model employs ResNet50 for extracting features from images in the first component, and uses decision trees, Support Vector Machine (SVM), and an ensemble algorithm for classification in the second component. The model was tested using three datasets: Real-World Masked Face Dataset (RMFD), Simulated Masked Face Dataset (SMFD), and Labeled Faces in the Wild (LFW). It achieved high testing accuracies, notably 99.64\% on RMFD, 99.49\% on SMFD, and 100\% on LFW. also B Wang et al.\cite{wang2021hybrid} outlines a two-stage hybrid machine learning approach for detecting mask-wearing in public spaces to reduce the spread of COVID-19. The first stage uses a pre-trained Faster R-CNN model combined with an InceptionV2 architecture to identify potential mask-wearing regions. The second stage employs a Broad Learning System (BLS) to verify these detections by differentiating actual mask-wearing from background elements. The method, tested on a new dataset comprising 7,804 images and 26,403 mask instances, demonstrates high accuracy, achieving 97.32\% in simple scenes and 91.13\% in complex scenes. And X Su et al.\cite{su2022face} integrates transfer learning and deep learning techniques to enhance accuracy and performance. Firstly, the face mask detection component employs Efficient-Yolov3 with EfficientNet as the backbone, using CIoU loss to improve detection precision and reduce computational load. Secondly, the classification component differentiates between 'qualified' masks (e.g., N95, disposable medical) and 'unqualified' masks (e.g., cotton, sponge masks) using MobileNet to overcome challenges associated with small datasets and overfitting.

BA Kumar\cite{kumar2023face} developed a face detection system capable of accurately identifying individuals whether they are wearing masks or not. This enhancement addresses the increased use of face masks in public due to the COVID-19 pandemic. The system leverages a modified Caffe-MobileNetV2 (CMNV2) architecture, where additional layers are integrated for better classification of masked and unmasked faces using fewer training parameters. The focus is on detecting facial features visible above the mask, such as the eyes, ears, nose, and forehead. The model demonstrated high accuracy, achieving 99.64\% on static photo images and similarly robust performance on real-time video
    
\begin{table*}[htbp]  
  \centering
  \begin{adjustbox}{width={\textwidth}}
  \begin{tabular}{|c|c|c|c|c|c|c|c|c|}
    \toprule
    \textbf{Model} & \textbf{Year} & \textbf{Dataset} & \textbf{Accuracy} & \textbf{Precision}  & \textbf{Recall} & \textbf{F1-score} & \textbf{AP} & \textbf{mAP} \\
    \midrule
    Efficient-Yolov3 \cite{su2022face} & 2022 & Face Mask Dataset &-&-&- & - & 98.18\% &  96.03\%\\
    \hline
    YOLOv3 \cite{singh2021face} & \multirow{2}{*}{2021} & \multirow{2}{*}{MAFA \& WIDER FACE} & &-&-&- & 55\% & - \\
    Faster R-CNN &  &    & - &  - & - &  - & 62\% &  -\\
    \hline
    \multirow{3}{*}{SE-YOLOv3 \cite{jiang2021real}} & \multirow{3}{*}{2021} & \multirow{3}{*}{PWMFD} & - &  - & - &  - & 73.7\% &  -\\
    &  &    & - & - & - & - & AP$_{50}$\ 99.5\% &  -\\
    &  &    & - & - & - & - & AP$_{75}$\ 88.7\% &  -\\
    \hline
    \makecell{Improved YOLO-v4\\ (CSPDarkNet53)} \cite{yu2021face} & 2021 & RMFD and Masked Face-Net & -& 93.6\% & 97.9\% & 95.7\% & 84.7\% &  -\\
    \hline
    YoloV5 \cite{ieamsaard2021deep} & 2021 & Kaggle and MakeM & 96.5\% & - &  - & - &  - &-\\
    \hline
    YOLOv5s-CA \cite{pham2023integration} & 2023 & \makecell{Kaggle + Created \\dataset from YouTube} &-& 95.9\% & 92.3\% & 94\% &- & mAP@0.5\ 96.8\% \\
    \hline
    \multirow{2}{*}{FMDYolo \cite{wu2022fmd}} & \multirow{2}{*}{2022} & \makecell{Kaggle(Face Mask \\Detection dataset)} & - & - & - & - & - & 66.4\% \\
    \cline{3-9}
    &  & VOC Mask  & - & - & - & - & - & 57.5\%\\
    \hline
    \multirow{2}{*}{AI-Yolo \cite{zhang2023novel}} & \multirow{2}{*}{2023} & Kaggle (WMD-1) & - & - & - & 89.3\% & - & 94.1\% \\
    \cline{3-9}
    &  &  Kaggle(WMD-2) & - & - & - & 78.6\% & - &    90.7\%\\
    \hline
    YOLOv8 \cite{tamang2023enhancing} & 2023 & \makecell{Face Mask Detection\\(FMD)}  &  -  &  95\%  &    95\% & - & - & mAP@0.5 \ 96\% \\    
    \bottomrule
  \end{tabular}
  \end{adjustbox}
    \caption{\label{tab:Face-Mask-Recognition-models2} Summary of YOLO-Based Face Mask Recognition Models}
\end{table*}
    
\subsubsection{YOLO (You Only Look Once)} is a real-time object detection system that recognises objects with a single forward pass through the neural network. This one-stage detector efficiently combines the tasks of object localization and identification, making it ideal for applications requiring rapid and accurate object detection, such as face mask detection. YOLO balances speed and precision, adapting to various scenarios where quick detection is crucial. S Singh et al.\cite{singh2021face} focus on face mask detection using two advanced deep learning models, YOLOv3 and Faster R-CNN, to monitor mask usage in public places during the COVID-19 pandemic. They developed a dataset of about 7500 images categorized into masked and unmasked faces, which they manually labeled and enhanced with bounding box annotations. This dataset includes various sources and is accessible online. Both models were implemented using Keras on TensorFlow and trained with transfer learning. The models detect faces in each frame and classify them as masked or unmasked, drawing colored bounding boxes (red or green) around the faces accordingly. Also X Jiang et al.\cite{jiang2021real}  introduces SE-YOLOv3, an enhanced version of the YOLOv3 object detection algorithm, optimized for real-time mask detection by integrating Squeeze and Excitation (SE) blocks into its architecture. This modification helps focus the network on important features by recalibrating channel-wise feature responses, significantly improving detection accuracy. SE-YOLOv3 also employs GIoULoss for precise bounding box regression and Focal Loss to handle class imbalance effectively. Additionally, the model uses advanced data augmentation techniques, including mixup, to enhance its generalization capabilities. \\

J Yu and W Zhang\cite{yu2021face} enhances the YOLO-v4 model for efficient and robust face mask recognition in complex environments, introducing an optimized CSPDarkNet53 backbone to minimize computational costs while enhancing model learning capabilities. Additionally, the adaptive image scaling and refined PANet structure augment semantic information processing. The proposed model is validated with a custom face mask dataset, achieving a mask recognition mAP of 98.3\%. J Ieamsaard et al.\cite{ieamsaard2021deep} investigates an effective face mask detection method using the YoloV5 deep learning model during the COVID-19 pandemic. By leveraging a dataset of 853 images categorized into "With\_Mask", "Without\_Mask", and "Incorrect\_Mask", the model was trained across different epochs (20, 50, 100, 300, and 500) to identify optimal performance. The results indicate that training the model for 300 epochs yields the highest accuracy at 96.5\%. This approach utilizes YoloV5's capabilities for real-time processing. Also, TN Pham et al.\cite{pham2023integration} developed two versions: YOLOv5s-CA, with the CA module before the SPPF layer, and YOLOv5s-C3CA, where CA replaces the C3 layers. Tested on a new dataset created from YouTube videos, YOLOv5s-CA achieved a mAP@0.5 of 96.8\%, outperforming baseline models and showing promising results for real-time applications in monitoring mask usage during the COVID-19 pandemic. The study also included an auto-labeling system to streamline the creation of training datasets.

\begin{figure*}[htbp] 
  \centering
  \includegraphics[width=\linewidth]{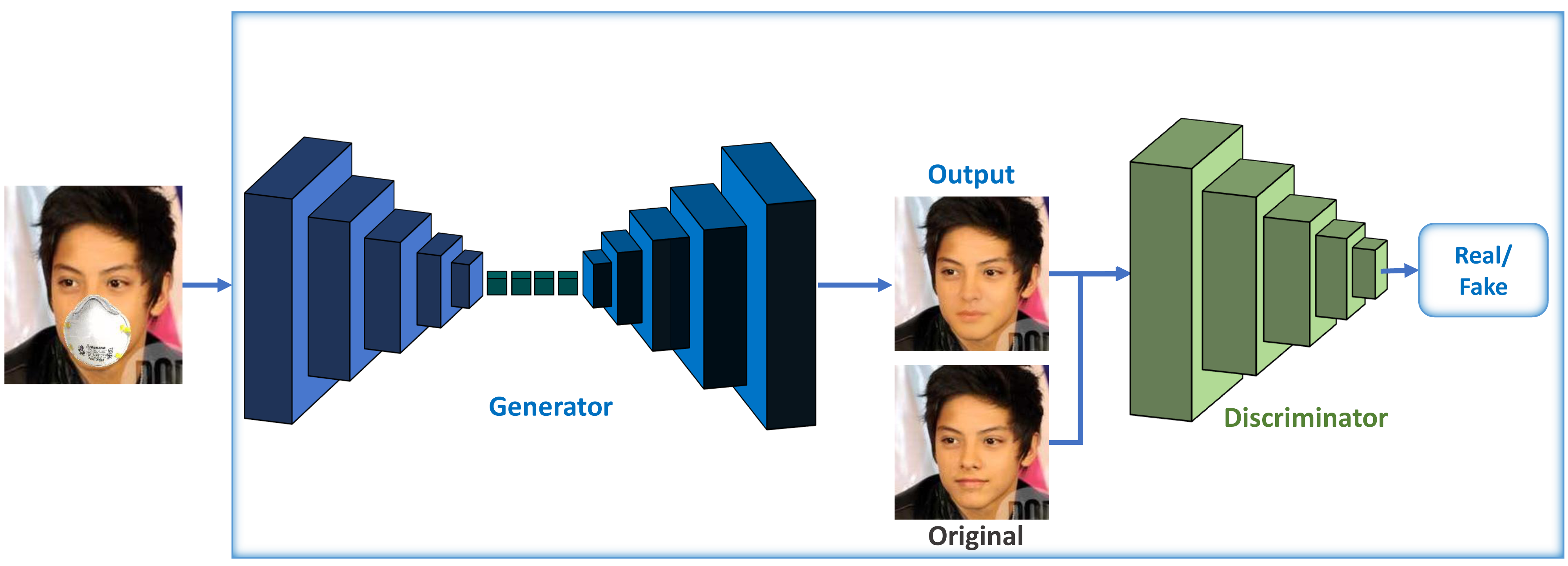}
  \caption{Overview of the GAN Network as an Example of FU-Net for Face Mask Removal.}
  \label{fig:2}
\end{figure*}

P Wu et al.\cite{wu2022fmd} proposed FMDYolo framework effectively detects whether individuals in public areas are wearing masks correctly, essential for preventing COVID-19 spread. It features the Im-Res2Net-101 as a backbone for deep feature extraction, combined with the En-PAN for robust feature fusion, improving model generalization and accuracy. The localization loss and Matrix NMS in training and inference stages enhance detection efficiency. And H Zhang et al.\cite{zhang2023novel} propose an enhanced object detection model named AI-Yolo, specifically designed for accurate face mask detection in complex real-world scenarios. The model integrates a novel attention mechanism through Selective Kernel (SK) modules, enhances feature representation using Spatial Pyramid Pooling (SPP), and promotes effective feature fusion across multiple scales with a Feature Fusion (FF) module. Additionally, it employs the Complete Intersection over Union (CIoU) loss function for improved localization accuracy. Also S Tamang et al.\cite{tamang2023enhancing} evaluate the YOLOv8 deep learning model for detecting and classifying face mask wearing conditions using the Face Mask Detector dataset. By employing transfer learning techniques, YOLOv8 demonstrated high accuracy in distinguishing between correctly worn masks, incorrectly worn masks, and no mask scenarios, outperforming the previous model, YOLOv5. The research highlights YOLOv8's enhancements in real-time object detection, making it suitable for applications requiring quick and reliable mask detection.

\subsection{Face Unmasking Approaches}
In this section, we explore recent progress in deep learning models designed for removing face masks, treating them as a specialized form of image inpainting, specifically focusing on 'object removal.' This technique offers promising opportunities not only for mask removal and restoring unmasked faces but also for applications like verification and identification systems. Figure \ref{fig:2} provides an overview of the GAN network as a representative example of the FU-Net. Additionally, Tables \ref{tab:FU-GAN-SOTA} and \ref{tab:FU-Diffusion-SOTA} outline popular models utilizing GAN and Diffusion methods, respectively.

\begin{table*}[htbp]
  \centering
  \begin{adjustbox}{width={\textwidth}}
  \begin{tabular}{|c|c|c|c|c|c|c|c|c|c|c|}
    \hline
    \textbf{Model} & \textbf{Year} & \textbf{Dataset} & \textbf{PSNR} & \textbf{SSIM} & \textbf{FID} & \textbf{NIQ} & \textbf{BRISQUE} & \textbf{MAE} & \textbf{$L_1$} & \textbf{$L_2$} \\
    \hline
    Context Encoders \cite{pathak2016context} & 2016 & Paris StreetView & 18.58 dB & - & - & - & - & - & 09.37\% &  1.96\%\\
    \hline
    \multirow{2}{*}{FFII-GatedCon \cite{yu2019free}} & \multirow{2}{*}{2019} & Places2(rectangular mask) & - & - & - & - & - & - &  8.6\% &  2.0\%\\
    \cline{3-10}
    &  & Places2(free-form mask) & - & - & - & - & - & - &  9.1\% &  1.6\%\\
    \hline
    EdgeConnect \cite{nazeri2019edgeconnect} & 2019 & Places2 & 21.75 & 0.823 & 8.16 & - & - & - &   3.86 &  - \\
    \hline
    PConv (N/B) \cite{liu2018image} & 2018 & Places2 & 18.21 / 19.04 & 0.468 / 0.484 & - & - & - & - &  6.45 / 5.72 & -\\
    \hline
    \multirow{3}{*}{R-MNet-0.4 \cite{jam2021r}} & \multirow{3}{*}{2021} & CelebA-HQ & 40.40 & 0.94 & 3.09 & - & - & 31.91 &  - &  -\\
    \cline{3-11}
    &   & Paris Street View & 39.55 & 0.91 & 17.64 & - & - & 33.81 &  - &  -\\
    \cline{3-11}
    &    & Places2 & 39.66 & 0.93 & 4.47 & - & - & 27.77 &  - &  -\\
    \hline
    MRGAN \cite{khan2019interactive} & 2019 & synthetic dataset & 29.91 dB & 0.937 & - & 3.548 & 29.97 & - &  - &  -\\
    \hline
    GFC (M5 \& Q5) \cite{li2017generative} & 2017 & CelebA & 19.5 & 0.784 & - & - & - & - &  - &  -\\
    \hline
    ERFGOFI (Mask) \cite{din2020effective} & 2020 & CelebA \& CelebA-HQ & 28.727 & 0.908 & - &  4.425 & 40.883 & - &  - &  -\\
    \hline
    GANMasker \cite{mahmoud2023ganmasker} & 2023 & CelebA & 30.96 & 0.95 & 16.34 & 4.46 &  19.27 & - &  - &  -\\
    GUMF \cite{din2020novel} & 2020 & CelebA &  26.19dB & 0.864 & 3.548 &  5.42 &   37.85 & - &  - &  -\\
    \hline
\end{tabular}
\end{adjustbox}
\caption{\label{tab:FU-GAN-SOTA} Summary of GAN-Based Face mask removal and image inpainting in general.}
\end{table*}

Most recent models in the field of object removal predominantly leverage GAN networks, a trend observed even before the emergence of COVID-19, as evidenced by works such as \cite{pathak2016context, iizuka2017globally, yu2019free, liu2018image, jam2021r, nazeri2019edgeconnect, li2017generative, khan2019interactive, din2020effective}. These methods are tailored for object removal tasks in general. However, there exists specific research dedicated to face mask removal, exemplified by works such as \cite{din2020novel, mahmoud2023ganmasker}, along with diffusion-based models \cite{lugmayr2022repaint, zhang2023towards, kawar2022denoising, wang2022zero}.

Upon delving into the realm of image inpainting methods, one encounters a diverse landscape of approaches pioneered by various researchers. Among the earliest methodologies stands the work of P. Deepak, K. Philipp, et al \cite{pathak2016context}, who introduced a convolutional neural network (CNN)-based technique employing context encoders to predict missing pixels. Building upon this foundation, Iizuka et al. \cite{iizuka2017globally} proposed a Generative Adversarial Network (GAN) framework equipped with two discriminators for comprehensive image completion. Similarly, Yu et al. \cite{yu2019free} put forth a gated convolution-based GAN tailored specifically for free-mask image inpainting. Nazeri et al. \cite{nazeri2019edgeconnect} devised a multi-stage approach involving an edge generator followed by image completion, facilitating precise inpainting. Additionally, Liu et al. \cite{liu2018image} contributed to the field with their work on free-mask inpainting, leveraging partial convolutions to exclusively consider valid pixels and dynamically update masks during the forward pass.

Another subset of methods focuses on face completion or the removal of objects from facial images. Jam et al. \cite{jam2021r} innovatively combined Wasserstein GAN with a Reverse Masking Network (R-MNet) for face inpainting and free-face mask completion. Similarly, Khan et al. \cite{khan2019interactive} leveraged a GAN-based network to effectively remove microphones from facial images. Li et al. \cite{li2017generative} devised a GAN architecture tailored specifically for generating missing facial components such as eyes, noses, and mouths. Further expanding the capabilities of inpainting, Ud Din et al. \cite{din2020effective} introduced a two-stage GAN framework enabling users to selectively remove various objects from facial images based on their preferences, with the flexibility to remove multiple objects through iterative application.

After the COVID-19 pandemic, considerable attention has been directed towards the development of techniques for face mask removal, encompassing both GAN-based and Diffusion-Based approaches. Among the GAN-based methodologies, Mahmoud \cite{mahmoud2023ganmasker} introduced a two-stage network architecture, initially focusing on face mask region detection to guide the subsequent inpainting stage. Additionally, Mahmoud enhanced their results by integrating Masked-unmasked region Fusion (MURF) mechanisms. Furthermore, Din et al. \cite{din2020novel} proposed a GAN-based network specifically designed for removing masks from facial images. Conversely, within the realm of diffusion models, Lugmayr et al. \cite{lugmayr2022repaint} introduced the RePaint method, which utilizes a DDPM \cite{ho2020denoising} foundation for image inpainting tasks. Similarly, Zhang et al. \cite{zhang2023towards} proposed the COPAINT method, enabling coherent inpainting of entire images without introducing mismatches. Broadly addressing image restoration, Kawar et al. \cite{kawar2022denoising} presented Denoising Diffusion Restoration Models (DDRM), offering an efficient, unsupervised posterior sampling approach for various image restoration tasks. In a related context, Wang et al. \cite{wang2022zero} devised the Denoising Diffusion Null-Space Model (DDNM), a novel zero-shot framework applicable to diverse linear image restoration problems, including image super-resolution, inpainting, colorization, compressed sensing, and deblurring.

\begin{table}[htbp]
  \begin{adjustbox}{width={0.5\textwidth}}
  \begin{tabular}{|c|c|c|c|c|}
    \hline
    \textbf{Model} & \textbf{Year} & \textbf{Dataset} & \textbf{Metric} & \textbf{Result}\\
    \hline
    \multirow{4}{*}{RePaint \cite{lugmayr2022repaint}} & \multirow{4}{*}{2022} & \multirow{2}{*}{CelebA-HQ} & LPIPS(Half) &  0.165\\
    &  &    & LPIPS(Expand) &   0.435\\
    \cline{3-5}
    &    &  \multirow{2}{*}{ImageNet} & LPIPS(Half) &  0.304\\
    &  &    & LPIPS(Expand) &  0.629\\
    \hline
    \multirow{4}{*}{COPAINT-TT \cite{zhang2023towards}} & \multirow{4}{*}{2023} & \multirow{2}{*}{CelebA-HQ} & LPIPS (Half) &   0.180\\
    &  &    & LPIPS (Expand) & 0.464\\
    \cline{3-5}
    &   &  \multirow{2}{*}{ImageNet} & LPIPS (Half) &   0.294\\
    &  &   & LPIPS (Expand) &  0.636\\
    \hline
    \multirow{4}{*}{DDRM-CC(SR) \cite{kawar2022denoising}} & \multirow{4}{*}{2022} & \multirow{4}{*}{ImageNet} & PSNR &   26.55\\
    &  &    & SSIM & 0.74\\
    &  &   &  KID & 6.56\\
    &  &   & LNFEs &  20\\
    \hline
    \multirow{6}{*}{ DDNM \cite{wang2022zero}} & \multirow{6}{*}{2022} & \multirow{3}{*}{ImageNet} & PSNR &   32.06\\
    &  &   & SSIM & 0.968\\
    &  & &  FID & 3.89\\
    \cline{3-5}
    &  &   \multirow{3}{*}{CelebA} & PSNR &    35.64\\
    &  &  & SSIM & 0.982\\
    &  &  &  FID & 4.54\\
    \hline
\end{tabular}
\end{adjustbox}
\caption{\label{tab:FU-Diffusion-SOTA} Summary of Diffusion-Based Face mask removal and image inpainting in general.}
\end{table}

\subsection{Masked Face Recognition Approaches}
In this subsection, we deeply into Deep Learning methodologies proposed to address the challenges faced by face recognition systems during the COVID-19 pandemic. The widespread use of masks has negatively affected the performance of traditional face recognition methods, encouraging authors to find novel approaches capable of effectively handling masked faces. Based on the important role of facial biometrics in various security systems and applications, it is important to address this issue by developing methods that perform robustly with both masked and unmasked faces. This subsection offers a comprehensive review of existing techniques for MFR, highlighting their diverse approaches and methodologies. The authors have pursued three distinct directions in masked face recognition, as illustrated in Figure \ref{fig:4}, delineated as follows:

\begin{figure*}[htbp] 
  \centering
  \includegraphics[width=\textwidth]{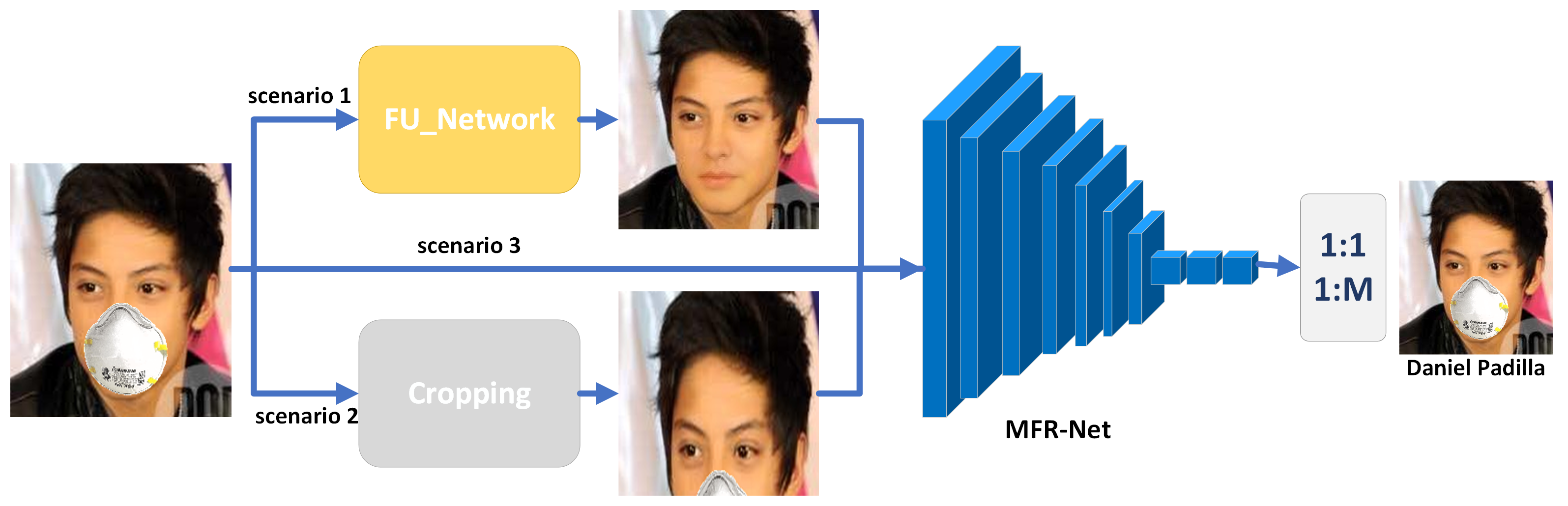}
  \caption{Three Directions in Masked Face Recognition (MFR): Face Restoration, Masked Region Discarding, and Deep Learning-Based Approaches.}
  \label{fig:4}
\end{figure*}

\subsubsection{Face Restoration} comprises two primary steps: initially, a model is employed for face unmasking to remove the mask and restore the hidden facial regions. Subsequently, another network is utilized to identify or verify the unmasked face. The primary objective within this category of the MFR task is to restore the face to its original, unmasked state. While the unmasking of faces and masked face recognition has traditionally been treated as distinct tasks, there are relatively few endeavors that amalgamate both within a single model. An example of such integration is evident in LTM \cite{li2020look}, depicted in the final row of Table \ref{tab:MRD-SOTA}.

LTM \cite{li2020look} proposes an innovative approach to enhance masked face recognition through the utilization of amodal completion mechanisms within an end-to-end de-occlusion distillation framework. This framework comprises two integral modules: the de-occlusion module and the distillation module. The de-occlusion module leverages a generative adversarial network to execute face completion, effectively recovering obscured facial features and resolving appearance ambiguities caused by masks. Meanwhile, the distillation module employs a pre-trained general face recognition model as a teacher, transferring its knowledge to train a student model on completed faces generated from extensive online synthesized datasets. Notably, the teacher's knowledge is encoded with structural relations among instances in various orders, serving as a vital posterior regularization to facilitate effective adaptation. Through this comprehensive approach, the paper demonstrates the successful distillation and transfer of knowledge, enabling robust identification of masked faces. Additionally, the framework's performance is evaluated across different occlusions, such as glasses and scarves. Notably, impressive accuracies of 98\% and 94.1\% are achieved, respectively, by employing GA \cite{yu2018generative} as an inpainting method.

\begin{table*}[htbp]
  \centering
  \begin{adjustbox}{width={\textwidth}}
  \begin{tabular}{|c|c|c|c|c|c|}
    \hline
    \textbf{Model} & \textbf{Year} & \textbf{Dataset} & \textbf{Accuracy} & \textbf{mAP} & \textbf{Rank-1}\\ 
    \hline
    \multirow{4}{*}{CA-MFR \cite{li2021cropping}} & \multirow{4}{*}{2021} & Masked-Webface (Case 1) & 91.525 & - & - \\
    \cline{3-6}
    &    & Masked-Webface (Case 2) & 86.853\% & -  & - \\
    \cline{3-6}
    &    & Masked-Webface (Case 3) & 81.421\% & - & - \\
    \cline{3-6}
    &    & Masked-Webface (Case 4) & 92.612\% & - & - \\
    \hline
    \multirow{2}{*}{Hariri \cite{hariri2022efficient}} & \multirow{2}{*}{2022} & RMFRD & 91.3\%  & - & -  \\
    \cline{3-6}
    &    & SMFRD &  88.9\%  & - & - \\
    \hline
    \makecell{IAMGAN-DCR\\(VGGFace2) \cite{geng2020masked}} & \multirow{2}{*}{2020} & \multirow{2}{*}{MFSR-REC} &  86.5\% & 42.7\% & 68.1\%\\
    \cline{1-1}
    \cline{4-6}
    \makecell{IAMGAN-DCR\\(CAISA-Webface) \cite{geng2020masked}} & & &  82.3\% & 37.5\% & 67.4\%\\ 
    \hline
    \makecell{UNMaskedArea-MFR \\ (Cosine Similarity) \cite{firdaus2022masked}} & 2022 & Custom (Indonesian people) Dataset & 98.88\% & - & - \\
    \hline
    \multirow{3}{*}{LPD \cite{ding2020masked}} & \multirow{3}{*}{2020} & MFV &  97.94\% & - & - \\
    \cline{3-6}
    &    &  MFI & 94.34\% & 49.08\% & -\\
    \cline{3-6}
    &    &  Synthesized LFW &  95.70\% & 75.92\% & -\\
    \cline{1-6}
    \hline
    \multirow{3}{*}{LTM \cite{li2020look}} & \multirow{3}{*}{2020} & LFW &  95.44\% & - & -\\
    \cline{3-6}
    &    & AR1&  98.0\% & - & -\\
    \cline{3-6}
    &    & AR2 &  94.1\% & - & -\\
    \hline
\end{tabular}
\end{adjustbox}
\caption{\label{tab:MRD-SOTA} Summary of Masked Region Discarding \& Face Restoration-Based For Masked Face Recognition.}
\end{table*}

\subsubsection{Masked Region Discarding} entails the removal of the masked area from the face image. This is achieved either by detecting the mask region and cropping out the masked portion or by using a predefined ratio to crop out a portion of the face that is typically unmasked. The remaining unmasked portion, usually containing facial features such as the eyes and forehead, is then utilized for training the recognition model. Table \ref{tab:MRD-SOTA} provides a concise overview of the methodologies associated with this approach.

One of these methods is Li et al. \cite{li2021cropping} explored the Occlusion Removal strategy by investigating various cropping ratios for the unmasked portion of the face. They incorporated an attention mechanism to determine the optimal cropping ratio, aiming to maximize accuracy. In their study, the optimal ratio was identified as 0.9L, where L represents the Euclidean distance between the eye keypoints. The authors conducted experiments across four scenarios. In the first scenario, the model was trained using fully masked images and tested on masked images, achieving an accuracy of 91.529\% with L = 0.9. The second and third scenarios involved training or testing with only one image masked, yielding accuracies of 86.853\% and 82.533\%, respectively, for L = 0.7. Finally, the fourth scenario adhered to the traditional face recognition approach, utilizing unmasked images for both training and testing purposes. Hariri \cite{hariri2022efficient} introduced a method that initially corrects the orientation of facial images and employs a cropping filter to isolate the unmasked areas. Feature extraction is conducted using pre-trained deep learning architectures including VGG-16, AlexNet, and ResNet-50. Leveraging pre-trained VGG-16 \cite{simonyan2014very}, AlexNet \cite{krizhevsky2017imagenet}, and ResNet-50 \cite{he2016deep} models for feature extraction in their masked face recognition approach, the effectiveness of these models in diverse image classification tasks has been well-documented, showcasing their ability to attain high recognition accuracy. Feature maps are extracted from the final convolutional layer of these models, followed by the application of the bag-of-features (BoF) \cite{passalis2017learning} methodology to quantize the feature vectors and create a condensed representation. The similarity between feature vectors and codewords is gauged using the Radial Basis Function (RBF) kernel. This approach demonstrates superior recognition performance compared to other state-of-the-art methods, as evidenced by experimental evaluations on the RMFRD \cite{Wang2020MaskedFaceRecognition} and SMFRD \cite{10036007} datasets, achieving accuracies of 91.3\% and 88.9\%, respectively.

Furthermore, G. Mengyue et al. \cite{geng2020masked} present a comprehensive strategy aimed at overcoming the challenges associated with MFR through the introduction of innovative methodologies and datasets. Initially, the authors introduce the MFSR dataset, which includes masked face images annotated with segmentation and a diverse collection of full-face images captured under various conditions. To enrich the training dataset, they propose the Identity Aware Mask GAN (IAMGAN), designed to synthesize masked face images from full-face counterparts, thereby enhancing the robustness of the dataset. Additionally, they introduce the Domain Constrained Ranking (DCR) loss to address intra-class variation, utilizing center-based cross-domain ranking to effectively align features between masked and full faces. Experimental findings on the MFSR dataset underscore the efficacy of the proposed approaches, underscoring their significance and contribution to the advancement of masked face recognition technologies. Fardause et al. \cite{firdaus2022masked} introduce an innovative training methodology tailored for MFR, leveraging partial face data to achieve heightened accuracy. The authors curate their dataset, consisting of videos capturing faces across a range of devices and backgrounds, featuring 125 subjects. Drawing from established methodologies, such as employing YOLOv4 \cite{bochkovskiy2020yolov4} for face detection, leveraging the pre-trained VGGFace model for feature extraction, and employing artificial neural networks for classification, the proposed system exhibits significant performance enhancements. While conventional training methods yielded a test accuracy of 79.58\%, the adoption of the proposed approach resulted in a notable improvement, achieving an impressive test accuracy of 99.53\%. This substantial performance boost underscores the effectiveness of employing a tailored training strategy for tasks related to masked face recognition.

Ding Feifei et al. \cite{ding2020masked} curated two datasets specifically tailored for MFR: MFV, containing 400 pairs of 200 identities for verification, and MFI, comprising 4,916 images representing 669 identities for identification. These datasets were meticulously developed to address the scarcity of available data and serve as robust benchmarks for evaluating MFR algorithms. To augment the training data and overcome dataset limitations, they introduced a sophisticated data augmentation technique capable of automatically generating synthetic masked face images from existing facial datasets. Additionally, the authors proposed a pioneering approach featuring a two-branch CNN architecture. In this architecture, the global branch focuses on discriminative global feature learning, while the partial branch is dedicated to latent part detection and discriminative partial feature learning. Leveraging the detected latent part, the model extracts discriminative features crucial for accurate recognition. Training the model involves utilizing both the original and synthetic training data, where images from both datasets are fed into the two-branch CNN network. Importantly, the parameters of the CNN in the two branches are shared, facilitating efficient feature learning and extraction.

\subsubsection{Deep Learning-based masked face approaches} center on leveraging deep learning techniques, often employing attention mechanisms to prioritize unmasked regions for feature extraction while attempting to mitigate the impact of the mask itself. Some authors opt to train models using a combined dataset of masked and unmasked faces, facilitating robustness to varying facial conditions. Unlike the previous methods, this approach does not require additional preprocessing steps such as face restoration or cropping of the upper face region. Table \ref{tab:DL-SOTA_1} presents a summary of methodologies employed within this paradigm.\\

\begin{table*}[htbp]
  \centering
  \begin{adjustbox}{width={\textwidth}}
  \begin{tabular}{|c|c|c|c|c|c|c|c|}    
    \hline
    \textbf{Model} & \textbf{Year} & \textbf{Dataset} & \textbf{Accuracy} & \textbf{Precision} & \textbf{Recall} & \textbf{F1-score} & \textbf{EER}\%\\
    \hline
    \multirow{5}{*}{MTArcFace \cite{montero2022boosting}} & \multirow{5}{*}{2022} & Masked-LFW  &  98.92\% & - & - & - & -\\
    \cline{3-8}
    & & Masked-CFP\_FF & 98.33\% & - & - & - & -\\
    \cline{3-8}
    & & Masked-CFP\_FP & 88.43\% & - & - & - & -\\
    \cline{3-8}
    & & Masked-AGEDB\_30 &   93.17\% & - & - & - & -\\
    \cline{3-8}
    & & MFR2 & 99.41\% & - & - & - & -\\
    \hline
    \multirow{4}{*}{MFCosface \cite{deng2021mfcosface}} & \multirow{4}{*}{2021} & LFW\_m & 99.33\% & - & - & - & -\\
    \cline{3-8}
    &  & CF\_m & 97.03\% & - & - & - & -\\
    \cline{3-8}
    &  & MFR2 & 98.50\% & - & - & - & -\\
    \cline{3-8}
    &  & RMFD & 92.15\% & - & - & - & -\\
    \hline
    \multirow{2}{*}{MFR-CDC \cite{wu2021masked}} & \multirow{2}{*}{2021} & SMFRD &  95.31\% & - & - & - & -\\
    \cline{3-8}
    & & RMFRD & 95.22\% & - & - & - & -\\
    \hline
    Deepmasknet \cite{ullah2022novel} & 2022 & MDMFR & 93.33\% & 93.00\% & 94.50\% &   93.74\% & -\\
    \hline
    \multirow{2}{*}{MFR-CNN\&LBP \cite{vu2022masked}} & \multirow{2}{*}{2022} & COMASK20 & - & 87\% &  87\% & 87\% & -\\
    \cline{3-8}
    &    & Essex dataset & - & 99\% & 97\% & 98\% & -\\
    \hline
    \multirow{3}{*}{\makecell{MFR-DML\& \\FaceMaskNet-21} \cite{golwalkar2022masked}} & \multirow{3}{*}{2022} & User dataset & 88.92\% & - & - & - & -\\
    \cline{3-8}
    & & RMFRD &  82.22\% & - & - & - & -\\
    \cline{3-8}
    & & User dataset &  88.186\% & - & - & - & -\\
    \hline
     RggNet \cite{kumar2021masked} & 2021 & custom dataset & 60.8\% &   77.7\% & 51.9 \% & - & -\\
    \hline
    \multirow{4}{*}{\makecell{Att-Based-MFR\\(CASIA-Webface\_m)} \cite{pann2022effective}} & \multirow{4}{*}{2022} & LFW\_m & 99.43\% & 99.30\% &  99.56\% &  99.43\% & -\\
    \cline{3-8}
    &  & AgeDB-30\_m & 95.86\% & 93.83\% &  97.82\% & 95.78\% & -\\
    \cline{3-8}
    &  & CFP-FP\_m & 97.74\% & 96.77\% &  98.69\% &  97.72\% & -\\
    \cline{3-8}
    &  & MFR2 & 96.75\% & 96.25\% &  97.22\% & 96.73\% & -\\  
    \hline
    \multirow{4}{*}{\makecell{Att-Based-MFR\\(VGG-Face2\_m)} \cite{pann2022effective}} & \multirow{4}{*}{2022} &LFW\_m & 99.41\%  & 99.26\% &  99.56\% &  99.40\% & -\\
    \cline{3-8}
    & & AgeDB-30\_m & 95.38\% & 93.10\%  &  98.11\% &   95.53\% & -\\
    \cline{3-8}
    &  & CFP-FP\_m & 96.98\% & 96.17\% &  98.40\% &  97.27\% & -\\
    \cline{3-8}
    &  & MFR2 & 99.00\% & 99.50\% &  98.45\% &  99.02\% & -\\    
    \hline
    Fine-Tuned MobileNet \cite{kocacinar2022real} & 2022 & MadFaRe (12 subjects) & 78.41\% & - & - & - & -\\
    \hline
    \multirow{4}{*}{MFCosface \cite{deng2021mfcosface}} & \multirow{4}{*}{2021} & LFW\_m & 99.33\% & - & - & - & -\\
    \cline{3-8}
    & & CF\_m & 97.03\% & - & - & - & -\\
    \cline{3-8}
    & & MFR2 &  98.50\% & - & - & - & -\\
    \cline{3-8}
    &  & RMFD & 92.15\% & - & - & - & -\\
    \hline
    \multirow{2}{*}{Lightweight CNN \cite{faruque975advanced}} & \multirow{2}{*}{2024} & HMFD (Frontal Image) & 98.00\% & 98.00\% & 97.00\% & 98.00\% & -\\
    \cline{3-8}
    & & HMFD (Lateral Image) & 79.00\% & 83.00\% &  80.00\% & 79.00\% & -\\
    \hline
    FaceNet + optimized SVM \cite{thanathamathee2023optimized} & 2023 & CASIA + LWF + User dataset & 99.912\% & - & - & - & -\\
    \hline
    YOLOv3 \cite{aswal2020single} & \multirow{3}{*}{2020} & \multirow{3}{*}{custom dataset} & 93\% & - & - & - & -\\
    \cline{4-8}
    YOLO-face + VGGFace2 & & & 96.8\% & - & - & - & -\\
    \cline{4-8}
    RetinaFace + VGGFace2 & & &  94.5\% & - & - & - & -\\
    \hline
    \multirow{4}{*}{ResNet-100-MR-MP(SRT)\cite{boutros2022self}}  & \multirow{4}{*}{2022} & MFR & - & - & - & - & 0.8270	\\
    \cline{3-8}
    & & MRF2 & - & - & - & - & 3.4416\\
    \cline{3-8}
    & & LFW & - & - & - & -  & 0.9667\\
    \cline{3-8}
    & & IJB-C & - & - & - & -  & 2.9197\\
    \hline
    \multirow{4}{*}{ResNet-50-MR-MP(SRT) \cite{boutros2022self}} & \multirow{4}{*}{2022} & MFR & - & - & - & -  & 1.1207\\
    \cline{3-8}
    & & MRF2 & - & - & - & - & 6.2578\\
    \cline{3-8}
    & & LFW & - & - & - & -  & 1.2333\\
    \cline{3-8}
    & & IJB-C & - & - & - & - & 3.0833\\
    \hline
    \multirow{4}{*}{MobileFaceNet-MR-MP(SRT) \cite{boutros2022self}} & \multirow{4}{*}{2022} & MFR & - & - & - & -  & 3.1866\\
    \cline{3-8}
    & & MRF2 & - & - & - & - & 7.8232\\
    \cline{3-8}
    & & LFW & - & - & - & -  & 2.2667\\
    \cline{3-8}
    & & IJB-C & - & - & - & - & 4.6837\\
    \hline
\end{tabular}
\end{adjustbox}
\caption{\label{tab:DL-SOTA_1} Summary of Deep Learning-Based Approaches For Masked Face Recognition.}
\end{table*}

Building upon the ArcFace architecture, Montero David et al \cite{montero2022boosting} introduced an end-to-end approach for training face recognition models, incorporating modifications to the backbone and loss computation processes. Additionally, they implemented data augmentation techniques to generate masked versions of the original dataset and dynamically combine them during training. By integrating the face recognition loss with the mask-usage loss, they devised a novel function termed Multi-Task ArcFace (MTArcFace). Experimental results demonstrated that their model serves as the baseline when utilizing masked faces, achieving a mean accuracy of 99.78\% in mask-usage classification, while maintaining comparable performance metrics on the original dataset.  On a parallel front, Deng Hongxia et al. \cite{deng2021mfcosface} proposed their masked-face recognition algorithm, leveraging large-margin cosine loss (MFCosface) to map feature samples in a space with reduced intra-class distance and expanded inter-class distance. They further developed a masked-face image generation algorithm based on the detection of key facial features, enabling the creation of corresponding masked-face images. To enhance their model's performance and prioritize unmasked regions, they introduced an Att-inception module combining the Inception-Resnet module and the convolutional block attention module. This integration heightened the significance of unoccluded areas in the feature map, amplifying their contribution to the identification process.  Additionally, Wu GuiLing \cite{wu2021masked} proposes a masked face recognition algorithm based on an attention mechanism for contactless delivery cabinets amid the COVID-19 pandemic. By leveraging locally constrained dictionary learning, dilated convolution, and attention mechanism neural networks, the algorithm aims to enhance recognition rates of masked face images. The model, validated on the RMFRD and SMFRD databases, demonstrates superior recognition performance. Furthermore, the algorithm addresses occlusion challenges by constructing subdictionaries for occlusion objects, effectively separating masks from faces. The network architecture incorporates dilated convolution for resolution enhancement and attention modules to guide model training and feature fusion. Overall, the proposed approach offers promising advancements in masked face recognition, crucial for ensuring the safety and efficiency of contactless delivery systems. Naeem Ullah et al. \cite{ullah2022novel} introduced the DeepMasknet model, a novel construction designed for face mask detection and masked facial recognition. Comprising 10 learned layers, the DeepMasknet model demonstrates effectiveness in both face mask detection and masked facial recognition tasks. Furthermore, the authors curated a large and diverse unified dataset, termed the Mask Detection and Masked Facial Recognition (MDMFR) dataset, to evaluate the performance of these methods comprehensively. Experimental results conducted across multiple datasets, including the challenging cross-dataset setting, highlight the superior performance of the DeepMasknet framework compared to contemporary models.

Vu Hoai Nam et al. \cite{vu2022masked} proposed a methodology that leverages a fusion of deep learning techniques and Local Binary Pattern (LBP) features for recognising masked faces. They employed RetinaFace, a face detector capable of handling faces of varying scales through joint extra-supervised and self-supervised multi-task learning, as an efficient encoder. Moreover, the authors extracted LBP features from specific regions of the masked face, including the eyes, forehead, and eyebrows, and integrated them with features learned from RetinaFace within a unified framework for masked face recognition. Additionally, they curated a dataset named COMASK20 comprising data from 300 subjects. Evaluation conducted on both the published Essex dataset and their self-collected COMASK20 dataset demonstrated notable improvements, with recognition results achieving an 87\% f1-score on COMASK20 and a 98\% f1-score on the Essex dataset. Golwalkar Rucha et al. \cite{golwalkar2022masked} introduced a robust masked face recognition system, leveraging the FaceMaskNet-21 deep learning network and employing deep metric learning techniques. Through the generation of 128-dimensional encodings, the system achieves precise recognition from static images, live video feeds, and video recordings in real-time. With testing accuracy reaching 88.92\% and execution times under 10 ms, the system demonstrates high efficiency suitable for a variety of applications. Its effectiveness in real-world scenarios, such as CCTV surveillance in public areas and access control in secure environments, positions it as a valuable asset for bolstering security measures amid the widespread adoption of face masks during the COVID-19 pandemic. Kumar Manoj and Mann Rachit \cite{kumar2021masked} delve into the implications of face masks on the efficacy of face recognition methods, with a specific emphasis on face identification employing deep learning frameworks. Drawing from a tailored dataset derived from VGGFace2 and augmented with masks for 65 subjects, the research scrutinizes the performance of prevalent pre-trained models like VGG16 and InceptionV3 after re-training on the masked dataset. Additionally, the study introduces a novel model termed RggNet, which capitalizes on a modified version of the ResNet architecture. This adaptation integrates supplementary layers within the shortcut paths of basic ResNet blocks, mirroring the structure of fundamental VGG blocks. This modification enables the model to effectively grasp an identity function, thereby fostering enhanced feature comprehension across layers. The proposed RggNet model architecture encompasses three sub-blocks organized akin to ResNet50v2, with customized identity blocks featuring convolution layers in lieu of direct shortcuts. Through meticulous experimental analysis, the study endeavors to offer valuable insights into bolstering masked face identification tasks amid the prevalent use of face masks in everyday contexts. Pann Vandet and Lee Hyo Jong \cite{pann2022effective} introduce an innovative approach to MFR utilizing deep learning methodologies, notably the convolutional block attention module (CBAM) and angular margin ArcFace loss. By prioritizing the extraction of critical facial features, particularly around the eyes, essential for MFR tasks, their method effectively addresses challenges posed by facial masks. To mitigate data scarcity, data augmentation techniques are employed to generate masked face images from traditional face recognition datasets. The refined ResNet-50 architecture acts as the backbone for feature extraction, augmented with CBAM to enhance efficiency in feature extraction. The resulting 512-dimensional face embeddings are optimized using the ArcFace loss function, leading to significant enhancements in MFR performance. Experimental findings corroborate the effectiveness of the proposed approach, underscoring its potential for practical applications within the realm of COVID-19 safety protocols. Kocacinar Busra et al. \cite{kocacinar2022real} presented a real-time masked detection service and mobile face recognition application aimed at identifying individuals who either do not wear masks or wear them incorrectly. Through the utilization of fine-tuned lightweight Convolutional Neural Networks (CNNs), the system achieves a validation accuracy of 90.40\% using face samples from 12 individuals. The proposed approach adopts a two-stage methodology: initially, a deep model discerns the mask status, categorizing individuals as masked, unmasked, or improperly masked. Subsequently, a face identification module employs traditional and eye-based recognition techniques to identify individuals. This system represents a significant advancement in masked face recognition, effectively addressing challenges associated with masks in digital environments.\\

Deng Hongxia et al. \cite{deng2021mfcosface} introduce MFCosface, an innovative algorithm tailored for masked-face recognition amid the challenges posed by the COVID-19 pandemic. To mitigate the shortage of masked-face data, the algorithm incorporates a novel masked-face image generation method that utilizes key facial features for realistic image synthesis. Departing from conventional triplet loss approaches, MFCosface employs a large margin cosine loss function, optimizing feature mapping to bolster inter-class discrimination. Moreover, an Att-inception module is introduced to prioritize unoccluded facial regions, essential for precise recognition. Experimental findings across diverse datasets underscore the algorithm's notable enhancement in masked-face recognition accuracy, presenting a promising solution for facial recognition in mask-wearing scenarios. Md Omar Faruque et al. \cite{faruque975advanced} propose a lightweight deep learning approach, leveraging the HSTU Masked Face Dataset (HMFD) and employing a customized CNN model to improve masked face identification. Integration of key techniques such as dropout, batch normalization, and depth-wise normalization optimizes model performance while minimizing complexity. In comparison to established deep learning models like VGG16 and MobileNet, the proposed model achieves a superior recognition accuracy of 97\%. The methodology encompasses dataset preprocessing, model creation, training, testing, and evaluation, ensuring robust performance in real-world scenarios. Transfer learning from pre-trained models such as VGG16 and VGG19, along with grid search for hyperparameter optimization, enhances model effectiveness. The architecture incorporates depthwise separable convolutions and carefully chosen layers to strike a balance between computational efficiency and accuracy, demonstrating exceptional performance even when facial features are partially obscured by masks. With an emphasis on simplicity and effectiveness, this lightweight CNN model offers a promising solution for recognising masked faces, contributing to public health and safety efforts during the pandemic.

Putthiporn Thanathamathee et al. \cite{faruque975advanced} conducted a study aimed at improving facial and masked facial recognition using deep learning and machine learning methods. Unlike previous research that often overlooked parameter optimization, this study employed a sophisticated approach. By integrating grid search, hyperparameter tuning, and nested cross-validation, significant progress was achieved. The SVM model, after hyperparameter tuning, achieved the highest accuracy of 99.912\%. Real-world testing confirmed the efficacy of the approach in accurately identifying individuals wearing masks. Through enhancements in model performance, generalization, and robustness, along with improved data utilization, this study offers promising prospects for strengthening security systems, especially in domains like public safety and healthcare. Vivek Aswal et al. \cite{thanathamathee2023optimized} introduce two methodologies for detecting and identifying masked faces using a single-camera setup. The first method employs a single-step process utilizing a pre-trained YOLO-face/YOLOv3 model. Conversely, the second approach involves a two-step process integrating RetinaFace for face localization and VGGFace2 for verification. Results from experiments conducted on a real-world dataset exhibit robust performance, with RetinaFace and VGGFace2 achieving impressive metrics. Specifically, they attained an overall accuracy of 92.7\%, a face detection accuracy of 98.1\%, and a face verification accuracy of 94.5\%. These methodologies incorporate advanced techniques such as anchor box selection, context attention modules, and transfer learning to enhance the efficiency and effectiveness of detecting masked faces and verifying identities. Fadi Boutros et al. \cite{boutros2022self} introduce an innovative method to improve masked face recognition performance by integrating the Embedding Unmasking Model (EUM) with established face recognition frameworks. Their approach incorporates the Self-restrained Triplet (SRT) loss function, enabling the EUM to generate embeddings closely resembling those of unmasked faces belonging to the same individuals. The SRT loss effectively addresses intra-class variation while maximizing inter-class variation, dynamically adjusting its learning objectives to ensure robust performance across various experimental scenarios. Leveraging fully connected neural networks (FCNN), the EUM architecture demonstrates adaptability to different input shapes, thereby enhancing its versatility. Rigorous evaluation of multiple face recognition models and datasets, including both real-world and synthetically generated masked face datasets, consistently reveals significant performance enhancements achieved by the proposed approach.

\section{Conclusion and Future Directions}
This survey paper has conducted a thorough investigation into recent progress and challenges within the realms of MFR, FMR, and FU. By examining various complexities, from the scarcity of datasets to the challenges posed by occlusion, we have provided insights into the intricate landscape of these tasks. Despite the hurdles identified, our survey has unveiled substantial advancements and innovations across these domains. Researchers and practitioners have made notable strides, from refining techniques for generating synthetic datasets to devising novel methods for gathering real mask faces, all aimed at addressing the multifaceted challenges of MFR, FMR, and FU.

Looking ahead, several critical areas warrant attention for future research and development in MFR and its related tasks. A primary focus should be on enhancing tools for generating synthetic datasets and intensifying efforts to collect authentic mask-face data. Overcoming dataset scarcity holds the key to bolstering the generalization capabilities and real-time performance of MFR systems. Concurrently, improving deep learning methods and exploring innovative ideas are essential to enhancing the outcomes of these tasks. Furthermore, there exists significant potential in integrating face unmasking with masked face recognition to enhance overall system accuracy. By incorporating face unmasking as a preprocessing step in MFR models, researchers can bolster accuracy and resilience, particularly in scenarios marked by varying levels of occlusion.

In summary, the journey of MFR, FMR, and FU continues to unfold. With sustained research endeavors and operational initiatives, we anticipate further progress and breakthroughs in these domains. By tackling current challenges head-on and embracing future directions, we can propel the field forward, unlocking new opportunities and applications along the way.

\section*{Funding}
This work was supported by the National Research Foundation of Korea(NRF) grant funded by the Korean government (Ministry of Science and ICT) (No. 2023R1A2C1006944, 50\%) and partly by Innovative Human Resource Development for Local Intellectualization program through the Institute of Information \& Communications Technology Planning \& Evaluation (IITP) grant funded by the Korea government (MSIT)
(IITP-2024-2020-0-01462, 50\%).

\bibliography{anthology,custom}
\bibliographystyle{unsrt}

\end{document}